\documentclass{article}

\usepackage[preprint]{neurips_2026}

\usepackage[utf8]{inputenc}
\usepackage[T1]{fontenc}
\usepackage{microtype}
\usepackage{graphicx}
\usepackage{subcaption}
\usepackage{booktabs}
\usepackage[table]{xcolor}
\usepackage{amsmath}
\usepackage{amssymb}
\usepackage{mathtools}
\usepackage{amsthm}
\usepackage{algorithm}
\usepackage{algorithmic}
\usepackage{makecell}
\usepackage{float}
\usepackage{hyperref}
\usepackage[capitalize,noabbrev]{cleveref}
\hypersetup{
  colorlinks=true,
  linkcolor=blue,
  citecolor=blue,
  urlcolor=blue
}

\definecolor{rankone}{HTML}{B7E4C7}
\definecolor{ranktwo}{HTML}{D8F3DC}
\definecolor{rankthree}{HTML}{F1F7C9}
\definecolor{rankfour}{HTML}{FFEFAD}
\definecolor{rankfive}{HTML}{FFD6A5}
\definecolor{ranksix}{HTML}{FFB4A2}

\theoremstyle{plain}

\theoremstyle{definition}

\theoremstyle{remark}

\def\showtags{1}
\usepackage{setspace}

\newcommand{\x}{\mathbf{x}}

\newcommand{\y}{\mathbf{y}}

\newcommand{\sympdnorm}{\mathcal{N}}
\newcommand{\pdnorm}[1]{\sympdnorm(#1)}

\usepackage[dvipsnames]{xcolor}
\usepackage{mathtools} % For \shortintertext
\newcommand{\bx}{\textbf{x}}
\newcommand{\btheta}{\bm{\theta}}
\newcommand{\bphi}{\bm{\phi}}

\usepackage{bm}

\newcommand{\bbeta}{\bm{\beta}}

\if\showtags1
\newcommand{\todo}[1]{{\textcolor{red}{\textbf{[TODO]} #1}}}
\newcommand{\done}[1]{{\textcolor{PineGreen}{\textbf{[Done]} #1}}}
\newcommand{\ttodo}[2]{{\textcolor{red}{\textbf{[TODO](#2)}: #1}}}

\newcommand{\toberemoved}[1]{{\textcolor{LimeGreen}{\textbf{[x]} #1}}}
\newcommand{\commentt}[1]{{\textcolor{Green}{\textbf{[Comment]} #1}}}

\else
\newcommand{\todo}[1]{}
\newcommand{\ttodo}[2]{}

\newcommand{\toberemoved}[1]{}
\newcommand{\commentt}[1]{}
\newcommand{\done}[1]{}
\fi

\usepackage{xspace} % Make sure the white space after these are not removed.

\title{Efficient Bayesian Deep Ensembles \\ via Analytic Predictive Inference}

\author{%
  Sina Aghaee Dabaghan Fard\textsuperscript{1} \quad
  Marie Maros\textsuperscript{1} \quad
  Jaesung Lee\textsuperscript{1}\thanks{Corresponding author: \texttt{j.lee@tamu.edu}}\\
  \textsuperscript{1}Wm Michael Barnes '64 Department of Industrial and Systems Engineering\\
  Texas A\&M University\\
  College Station, TX, USA
}

\begin{document}

\maketitle

\begin{abstract}
We introduce an efficient Bayesian deep ensemble method for predictive regression designed to enhance interpretability while maintaining competitive predictive performance and computational efficiency. Our method combines the statistical rigor of Bayesian inference with the scalability of deep ensembles, providing calibrated uncertainty estimates that enable its use not only for standalone prediction but also as a component within broader learning systems. To achieve these goals, our work relies on three key design components:
\textbf{(i) low-dimensional ensemble representation:} predictions are expressed as a combination of a small number of trained neural predictors, enabling scalable inference whose cost depends on ensemble size rather than dataset size;
\textbf{(ii) closed-form Bayesian aggregation:} ensemble predictions are combined using Bayesian linear regression, yielding interpretable posterior weights and calibrated uncertainty without approximate inference; and
\textbf{(iii) Independent ensemble training:} multiple neural networks are trained separately, producing diverse predictive representations that improve robustness and uncertainty calibration.
Empirical results on standard regression benchmarks demonstrate that the proposed approach achieves competitive predictive performance while maintaining reliable uncertainty estimates across settings.
\end{abstract}

\begin{figure}[h]
    \centering
\includegraphics[width=0.65\linewidth]{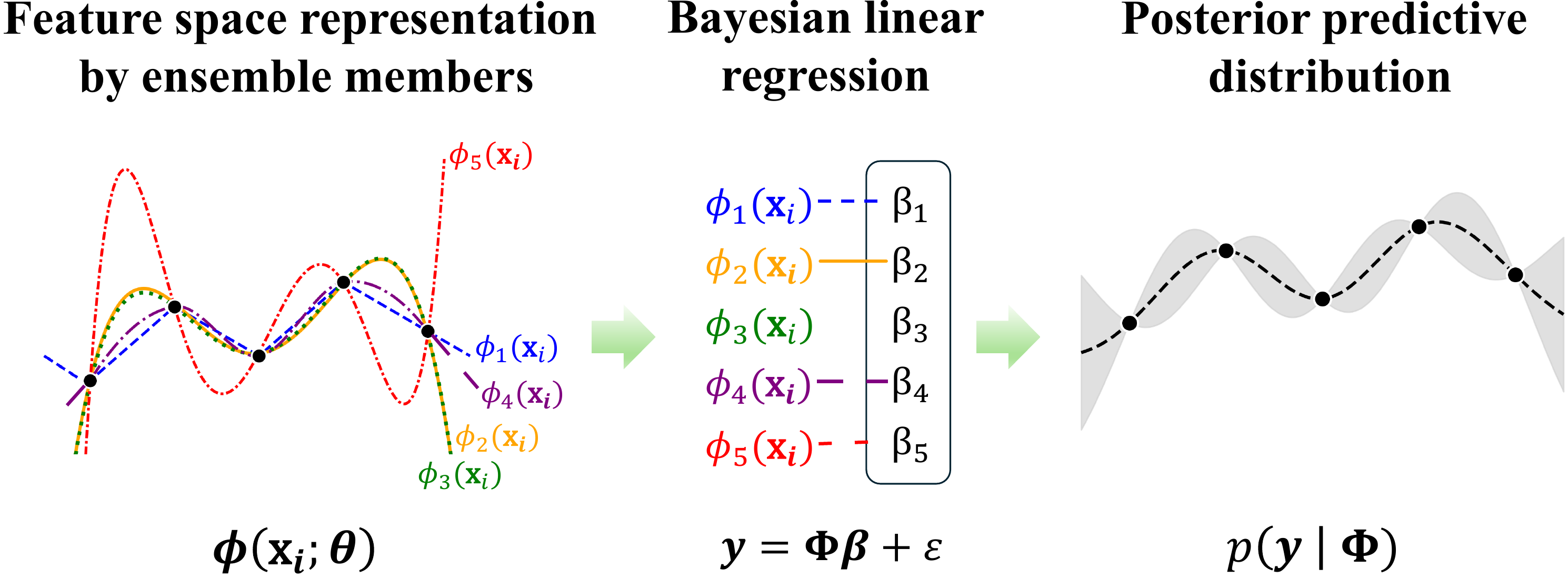}
    \caption{
    \textbf{Proposed framework.}
    An ensemble of neural networks induces a feature space representation
    $\{\phi_h(\bx_i;\theta_h)\}_{h=1}^H$.
    A Bayesian linear model is placed on top of these ensemble-induced features, and the
    linear weights $\boldsymbol{\beta}$ and noise variance $\sigma^2_{\varepsilon}$ are integrated out with respect to their posteriors,
    yielding a closed-form posterior predictive distribution over targets $\y$.    \label{fig:bdkn_overview}
}
    \end{figure}

\section{Introduction}

We consider predictive regression models in settings where reliable uncertainty estimates are required alongside \textbf{(i)} computational efficiency, \textbf{(ii)} strong predictive performance, and \textbf{(iii)} interpretability. This requirement is particularly important in safety-critical and risk-sensitive applications \citep{filos2019systematic}, where deep neural networks (NNs) are known to produce poorly calibrated predictions that can be overconfident \citep{guo2017calibration,lakshminarayanan2017simple,watson2021latent}, especially under distribution shift. Moreover, regression models with quantified uncertainty serve as core components in broader learning systems, including reinforcement learning \cite{sutton1998reinforcement, arulkumaran2017deep, levine2020offline, deisenroth2011pilco} and physics-informed neural networks \cite{raissi2019physics}. Goals \textbf{(i)}--\textbf{(iii)} are inherently in tension. In what follows, we examine several widely used paradigms that address these requirements to varying degrees while exposing important trade-offs.

 In terms of predictive performance and principled uncertainty estimation, Gaussian Processes (GPs) provide a canonical Bayesian framework for nonlinear regression with exact posterior inference \citep{williams2006gaussian}. However, their cubic computational complexity and sensitivity to kernel choice fundamentally limit scalability, particularly in high-dimensional settings where distance-based kernels suffer from the curse of dimensionality. Deep Kernel Learning (DKL) partially mitigates this limitation by learning data-dependent covariance functions using neural networks, but exact inference remains computationally intensive, and scalable implementations typically rely on approximations \citep{wilson2016deep}.

Fully Bayesian neural networks (BNNs) represent a different point in the performance–computation trade-off by placing probability distributions over network parameters \citep{neal2012bayesian, wilson2020bayesian}. While they provide a flexible probabilistic framework for modeling uncertainty, exact inference is generally intractable, and practical implementations rely on approximate inference methods such as mean-field variational inference \citep{graves2011practical}. These approximations can impose restrictive assumptions on the posterior distribution, which may lead to miscalibrated uncertainty estimates \citep{foong2020expressiveness, wenzel2020good}.

Bayesian last-layer (BLL) methods address the computational limitations of fully Bayesian neural networks by restricting Bayesian inference to the final linear layer of a neural network. This design enables efficient inference while retaining a probabilistic interpretation of predictions \citep{lazaro2010marginalized}. However, because uncertainty is placed only on a single learned representation, the representation itself is treated as fixed. As a result, uncertainty estimates can become overconfident when the learned representation fails to capture variability in the data, particularly outside the training distribution \citep{watson2021latent}. Extensions such as functional regularization have been proposed to mitigate this limitation, but they introduce additional computational overhead without fundamentally addressing the reliance on a single representation \citep{watson2021latent}.

Deep ensembles (DE) provide a simple and scalable approach to uncertainty quantification by training multiple neural networks independently and uniformly averaging their predictive distributions \citep{lakshminarayanan2017simple}. These methods typically yield strong empirical performance while keeping computational costs moderate. However, they lack an explicit probabilistic interpretation, as uncertainty is quantified implicitly through model diversity rather than through a well-defined probability measure over functions. Moreover, uniform averaging is non-diagnostic: it does not provide probabilistic evidence for which ensemble members are supported by the data or how individual members contribute to predictions. As a result, performance can degrade when member quality varies, increasing the risk of overconfident predictions.

Taken together, these approaches illustrate a recurring pattern: methods that provide principled uncertainty estimation often incur substantial computational cost or restrict model flexibility, while scalable approaches typically rely on heuristic aggregation rules that offer limited interpretability. Designing models that simultaneously achieve computational efficiency, strong predictive performance, and interpretable uncertainty estimation, therefore, remains an open challenge.

\subsection{Main contributions}

In this work, we propose \emph{Bayesian Deep Kernel Networks}, a regression framework that combines independently trained neural networks with closed-form Bayesian aggregation. The method trains $H$ neural networks on a dataset $\mathcal{D} = \{(\mathbf{x}_i,y_i)\}_{i=1}^N$, producing multiple learned feature representations that serve as basis functions. A Bayesian linear regression layer is then applied to these ensemble outputs, yielding an analytic posterior over aggregation weights and a closed-form posterior predictive distribution (Figure~\ref{fig:bdkn_overview}). This design enables principled uncertainty estimation while retaining the computational efficiency and predictive performance of standard deep ensembles.

The framework leverages the scalability of deep ensembles while replacing uniform averaging with Bayesian aggregation. By performing exact inference at the aggregation stage, the method produces interpretable posterior weights and calibrated uncertainty estimates without relying on approximate inference. Taken together, this design addresses the competing requirements \textbf{(i)}--\textbf{(iii)}. We summarize the resulting contributions below.

\textbf{Scalable Bayesian aggregation with learned neural predictors and finite-rank GP interpretation.}
We aggregate independently trained neural networks using closed-form Bayesian linear regression, yielding an analytic posterior predictive distribution with computational cost comparable to standard deep ensembles. The resulting model can be interpreted as a finite-rank Gaussian Process whose basis functions are learned directly through ensemble training, enabling principled Bayesian aggregation without kernel design or cubic-cost GP inference.

{\textbf{Separation of representation learning and probabilistic inference.}
Our framework maintains the standard deep-ensemble training pipeline and performs Bayesian inference only at the aggregation stage. Unlike Bayesian last-layer (BLL) methods that place uncertainty on a single latent representation, our method defines uncertainty directly over ensemble-member contributions. This separation maintains implementation simplicity while providing interpretable probabilistic weights for individual predictors.

{\textbf{Posterior-driven aggregation robust to heterogeneous ensemble member quality.}
Instead of uniform averaging, the aggregation weights are inferred from the posterior distribution. Ensemble members weakly supported by the data are automatically downweighted, improving robustness when model quality varies due to optimization variability or training conditions. This robustness is achieved without manual model selection or additional tuning heuristics.

To evaluate the practical implications of these design choices and their impact on computational efficiency, predictive performance, and interpretability, we conduct experiments on standard UCI regression benchmarks across a range of architectural and optimization settings.

\section{Bayesian deep kernel networks}
\label{sec:bdkn}

In this section, we introduce our framework while highlighting each improvement in comparison to existing methods. We start by introducing BLL and proceed to build upon it.

\subsection{Learning the feature space representation by ensemble members}

Let $\mathcal{D} = \{(\bx_i, y_i)\}_{i=1}^N$ denote the dataset,
where $\bx_i \in \mathbb{R}^P$ are input vectors and $y_i \in \mathbb{R}$ are scalar
responses. We consider scalar responses throughout this work for simplicity and without loss of generality. Denote by $\phi(\cdot;\boldsymbol{\theta}): \mathbb{R}^{P} \to \mathbb{R}^H$ a feature map with parameter $\boldsymbol{\theta}.$ In BLLs the latent function $f$ is modeled as
\begin{align}
% $$
y_i = f(\bx_i;\btheta)=\bphi(\bx_i;\btheta)^\top \bbeta+ \varepsilon_i
,\quad
    % where $
    \varepsilon_i \sim \mathcal{N}(0, \sigma^2_{\varepsilon})
    % $
\end{align}
where $\bbeta\sim\pdnorm{\mu_{\beta},\Sigma_{\beta}}$ denotes the feature weights.

Stacking the responses in the vector $\mathbb{R}^N$ yields the compact representation
\begin{align}\label{eq:lin_reg}
    \mathbf{y} = \boldsymbol{\Phi}(\mathbf{x};\boldsymbol{\theta})\boldsymbol{\beta} + \boldsymbol{\varepsilon},
\end{align}
where $\boldsymbol{\Phi}(\mathbf{x};\boldsymbol{\theta}) = [\boldsymbol{\phi}(\mathbf{x}_1;\boldsymbol{\theta}),\hdots,\boldsymbol{\phi}(\mathbf{x}_N;\boldsymbol{\theta})\boldsymbol]^{\top} \in \mathbb{R}^{N\times H}.$ Then, training is performed by finding $\boldsymbol{\theta},\,\sigma_{\varepsilon}^2,$ and $\boldsymbol{\Sigma}_{\beta}$ which requires maximizing an exact or approximate marginal likelihood and $\bbeta$ is obtained through its posterior $p(\bbeta \mid \mathcal{D}, \theta)$ in closed form \cite{watson2021latent}. Observe that the feature map $\phi(\mathbf{x};\boldsymbol{\theta})$ is learned once, and then treated as fixed during Bayesian inference. This implies that the variability employed to capture uncertainty stems only from the posterior over the weights $\boldsymbol{\beta}$ with the representation itself remaining deterministic. Therefore, uncertainty is restricted only to linear combinations within a single learned feature space.

In order to improve uncertainty behavior in out-of-distribution regions, we combine ideas from DE-style representation with a BLL to obtain better calibrated uncertainty. Unlike standard BLL, which trains a single neural network to construct $\boldsymbol{\Phi}.$ Thus, we define a collection of feature maps $\{\phi_h(\cdot;\boldsymbol{\theta}_h)\}_{h=1}^H,$ where each element is an individual map parameterized by its own $\boldsymbol{\theta}_h,$ yielding 
\begin{align*}
    \boldsymbol{\Phi}(\mathbf{x};\boldsymbol{\theta}) = [\phi_1(\mathbf{x}_1;\boldsymbol{\theta}_1),\hdots, \phi_h(\mathbf{x}_h;\boldsymbol{\theta}_h)]^{\top}
\end{align*}
in \eqref{eq:lin_reg}. Each feature map of the $H$ feature maps is implemented by an independently trained neural network.~Training is performed by maximizing the Gaussian likelihood of $y_i \sim \mathcal{N}(\phi_h(\mathbf{x}_i;\theta_h),\sigma^2_h(\mathbf{x}_i;\theta_h))$} where both the mean function $\phi_h(\cdot)$ and the variance function $\sigma_h^2(\cdot)$ are outputs of the network.  Note that $\sigma^2_h(\cdot)$ is used only during training to induce a heteroscedastic objective with the goal of preventing all networks from fitting the data with the same smoothness profile. It reweights the contribution of each training point to the loss: points with larger predicted variance incur a smaller penalty, while those with smaller variance are fitted more tightly. This sample-dependent reweighting leads to heterogeneous learned feature maps across independently trained ensemble members.

The independent training yields the following benefits for our model. First, feature maps can be trained fully in parallel, allowing the ensemble size to scale without increasing training time beyond that of a single neural network. Second, each feature map represents a complete predictive model of the data, rather than a latent basis.  In this way, the collection of independently learned feature maps captures multiple distinct functional representations of the data, which is particularly relevant when data are scarce compared to model size.

\subsection{Bayesian inference in the induced function space}

Once $\boldsymbol{\Phi}$ has been obtained, the posterior distribution over $\boldsymbol{\beta} \in \mathbb{R}^H$ is inferred via Bayesian linear regression in the function space \eqref{eq:lin_reg}.}

% \subsubsection*{Conjugate Prior}

We place a conjugate prior over the linear coefficients $\boldsymbol{\beta}$ and
the observation-noise variance $\sigma^2_{\varepsilon}$. We adopt Jeffreys' prior for the noise variance to improve generalizability and uncertainty quantification,
\begin{equation}
p(\sigma^2_{\varepsilon}) \propto \frac{1}{\sigma^2_{\varepsilon}},
\end{equation}
which is invariant under reparameterization and introduces no additional scale hyperparameters (See Appendix~\ref{subsub:jeffreys} for derivation). 
% \todo{think about whether mentioning invariant}
% 
Conditioned on $\sigma^2_{\varepsilon}$, the regression coefficients are assigned a Gaussian
prior,
\begin{equation}
\boldsymbol{\beta} \mid \sigma^2_{\varepsilon} \sim
\mathcal{N}\!\big(\boldsymbol{\mu}_0,\; \sigma^2_{\varepsilon} \boldsymbol{\Lambda}_0^{-1}\big),
\end{equation}
where $\boldsymbol{\Lambda}_0 \in \mathbb{R}^{H \times H}$ is positive definite.

% \commentt{I wrote the following to mention that Jeffreys’ prior is not the only option, but I think it is redundant since we already discuss this and mentioned that Jeffreys’ prior is a special case. Maybe instead, at the end of the paragraph, I can simply add: “which highlights the advantage of Jeffreys’ prior over alternative choices such as the normal-inverse-gamma prior.”}

% \textcolor{red}{An inverse-gamma prior on $\sigma_\epsilon^2$ could also be employed while maintaining conjugacy. We adopt Jeffreys' prior instead, as it avoids introducing additional hyperparameters that require tuning while still yielding a proper posterior predictive distribution.} 
This prior specification can be viewed as a special case of the
Normal-Inverse-Gamma prior, where
% $\boldsymbol{\beta}\mid\sigma^2_{\varepsilon} \sim \mathcal{N}(\boldsymbol{\mu}_0,\sigma^2_{\varepsilon}\boldsymbol{\Lambda}_0^{-1})$
% and 
$\sigma^2_{\varepsilon} \sim \mathrm{Inv\text{-}Gamma}(a_0,b_0)$, obtained in the
limit $a_0 \to 0$ and $b_0 \to 0$. 
% Although Jeffreys' prior is improper itself, meaning it does not integrate to one and therefore is not a probability distribution, the resulting posterior and posterior predictive distributions are proper and have closed-form expressions. 
Jeffreys' prior is a common choice when no prior information on variance is provided \citep{jeffreys1946invariant, gelman1995bayesian}, while maintaining analytical
tractability and eliminating the need to estimate additional hyperparameters for
the variance.

\noindent\textbf{Posterior updates.}
Given the design matrix $\boldsymbol{\Phi}$ and targets
$\mathbf{y}$, conjugacy is maintained and the posterior factorizes as \(
p(\boldsymbol{\beta},\sigma^2_{\varepsilon}\mid \mathbf{y},\boldsymbol{\Phi})
= p(\boldsymbol{\beta}\mid\sigma^2_{\varepsilon},\mathbf{y},\boldsymbol{\Phi})\;
  p(\sigma^2_{\varepsilon}\mid \mathbf{y},\boldsymbol{\Phi}).
\)
The conditional posterior over coefficients is Gaussian,
\(
\boldsymbol{\beta}\mid\sigma^2_{\varepsilon},\mathbf{y},\boldsymbol{\Phi}
\sim \mathcal{N}\!\big(\boldsymbol{\mu}_n,\; \sigma^2_{\varepsilon}\boldsymbol{\Lambda}_n^{-1}\big),
\)
with \(
\boldsymbol{\mu}_n = \boldsymbol{\Lambda}_n^{-1}
\big(\boldsymbol{\Lambda}_0\boldsymbol{\mu}_0 + \boldsymbol{\Phi}^\top \mathbf{y}\big)\)  and
\(
\boldsymbol{\Lambda}_n = \boldsymbol{\Lambda}_0 + \boldsymbol{\Phi}^\top\boldsymbol{\Phi}.
\)
Under the Jeffreys prior on $\sigma^2_{\varepsilon}$, the marginal posterior over $\sigma^2_{\varepsilon}$ is \(
\sigma^2_{\varepsilon}\mid \mathbf{y},\boldsymbol{\Phi}\sim \mathrm{Inv\text{-}Gamma}(a_n,b_n),
\)
where \(
a_n = \frac{N}{2}\) and \(
b_n = \frac{1}{2}\!\left(
\mathbf{y}^\top\mathbf{y}
+ \boldsymbol{\mu}_0^\top\boldsymbol{\Lambda}_0\boldsymbol{\mu}_0
- \boldsymbol{\mu}_n^\top\boldsymbol{\Lambda}_n\boldsymbol{\mu}_n
\right).
\)

\noindent\textbf{Posterior predictive.}
For a new input with feature vector $\boldsymbol{\phi}_* = \boldsymbol{\phi}(\mathbf{x}_\star)$, the predictive
distribution integrates out both $\boldsymbol{\beta}$ and $\sigma^2_{\varepsilon}$,
yielding a Student-$t$ distribution,
\begin{equation}
\begin{aligned}
p(y_*\mid \boldsymbol{\phi}_*,\mathbf{y},\boldsymbol{\Phi})
= \mathrm{St}\!\Big(
y_*;\;
\nu = 2a_n,\;
m = \boldsymbol{\phi}_*^\top\boldsymbol{\mu}_n, 
s^2 = \frac{b_n}{a_n}
\big(1+\boldsymbol{\phi}_*^\top\boldsymbol{\Lambda}_n^{-1}\boldsymbol{\phi}_*\big)
\Big).
\label{eq:posterior prediction}
\end{aligned}
\end{equation}

% In \eqref{eq:posterior prediction}, $\bm{\mu}_0$ and $\bm{\Lambda}_0$ denote hyperparameters that must be specified. In the Bayesian linear regression and BLL literature, these hyperparameters are commonly set to $\bm{\mu}_0 = \mathbf{0}$ and $\bm{\Lambda}_0 = \sigma^2_{\beta} \mathbf{I}$, with $\sigma^2_{\beta}$ estimated via empirical Bayes, which we also adopt \citep{watson2021latent}. See Appendix~\ref{subsub:beta_isotropic} for more details.
% % 
% % \noindent\textbf{Marginal likelihood.}
% % 
% The hyperparameter $\bm{\Lambda}_0$, specifically $\sigma^2_{\beta}$, is optimized by an empirical Bayes estimate, maximizing the marginal likelihood in a closed form. Algorithm~\ref{alg:bdkn} in Appendix~\ref{subsec:bdkn_algo} outlines the main steps of BDKN.

In \eqref{eq:posterior prediction}, $\bm{\mu}_0$ and $\bm{\Lambda}_0$ denote hyperparameters that must be specified. In the Bayesian linear regression and BLL literature, these are commonly set to $\bm{\mu}_0 = \mathbf{0}$ and $\bm{\Lambda}_0 = \sigma^2_{\beta}\mathbf{I}$, where $\sigma^2_{\beta}$ is estimated via empirical Bayes, which we also adopt \citep{watson2021latent}. Specifically, $\sigma^2_{\beta}$ is obtained by maximizing the closed-form marginal likelihood. See Appendix~\ref{subsub:beta_isotropic} and Appendix~\ref{subsec:bdkn_algo} for more details. Algorithm~\ref{alg:bdkn} in Appendix~\ref{subsec:bdkn_algo} outlines the main steps of BDKN.

\subsection{Connection to Bayesian model}

Our method has a direct interpretation in terms of Gaussian Processes. Since $f(\bx)$ is an affine transformation of a Gaussian random vector, it follows
that, conditional on $\sigma^2_{\varepsilon}$, $f(\bx)$
is a Gaussian Process,
\[
f(\cdot)\mid\sigma^2_{\varepsilon} \sim \mathcal{GP}\!\big(m(\cdot),k(\cdot,\cdot\mid\sigma^2_{\varepsilon})\big).
\]
where the mean function is
\[
m(\bx)=\mathbb{E}[f(\bx)]=\boldsymbol{\phi}(\mathbf{\bx})^\top\boldsymbol{\mu}_0,
\]
and the covariance function follows as
\begin{align*}
k(\bx,\bx'\mid\sigma^2_{\varepsilon})
=\mathrm{Cov}\!\big(f(\bx),f(\bx')\mid\sigma^2_{\varepsilon}\big) =\sigma^2_{\varepsilon}\,\boldsymbol{\phi}(\mathbf{x})^\top\Lambda_0^{-1}\boldsymbol{\phi}(\mathbf{x'}).
\end{align*}

% Letting $\Lambda_0^{-1/2}$ denote symmetric square root of $\Lambda_0^{-1}$, this kernel has the inner-product form
% \[
% k(x,x'\mid\sigma^2_{\varepsilon})
% = \sigma^2_{\varepsilon}\big(\Lambda_0^{-1/2}\boldsymbol{\phi}(\mathbf{x})\big)^\top
%     \big(\Lambda_0^{-1/2}\boldsymbol{\phi}(\mathbf{x'})\big).
% \]
With $\Lambda_0 = \sigma_{\beta}^{-2} I$, the kernel reduces to \(
k(\bx,\bx'\mid\sigma^2_{\varepsilon}) = \sigma^2_{\varepsilon}\sigma_{\beta}^2 \sum_{h=1}^H \phi_h(\bx) \phi_h(\bx'),
\)
which corresponds to an isotropic Gaussian prior over the coefficients. See Appendix~\ref{subsub:gp_connection} and Appendix~\ref{subsub:beta_isotropic} for more details and derivations. Conditioned on the learned feature maps $\{\phi_h\}_{h=1}^H$, the hypothesis space of the induced Gaussian process is the finite-dimensional linear span \(
\mathcal{H}_H = \mathrm{span}\{\phi_1,\dots,\phi_H\}.
\) Marginalizing the noise variance $\sigma^2_{\varepsilon}$ under the Jeffreys prior yields a Student-$t$ process \citep{shah2014student} with the same finite-dimensional support $\mathcal{H}_H$, which is consistent
with the closed-form Student-$t$ posterior predictive distribution presented in~\eqref{eq:posterior prediction}.

\subsection{Computational complexity}

Computationally, BDKN has nearly the same practical cost as a standard deep ensemble. A DE trains \(H\) independent neural networks and aggregates their predictions by uniform averaging, so its dominant cost is the cost of training the \(H\) ensemble members. BDKN uses the same ensemble training stage, but replaces uniform averaging with a Bayesian aggregation layer over the ensemble-member outputs. Given the ensemble-induced design matrix \(\boldsymbol{\Phi}\in\mathbb{R}^{N\times H}\), this aggregation requires forming an \(H\times H\) matrix and solving a small \(H\)-dimensional Bayesian linear regression problem, giving an additional cost of \(O(NH^2 + H^3)\) time and \(O(H^2)\) memory. The \(O(NH^2)\) term comes from forming \(\boldsymbol{\Phi}^\top\boldsymbol{\Phi}\), which sums pairwise interactions between ensemble members over all \(N\) training points, while the \(O(H^3)\) term comes from solving the resulting \(H\times H\) linear system. Since H is small and fixed in our proposed method, this additional cost is negligible relative to neural network training and remains linear in \(N\). 

\section{Experiments}
\label{sec:experiments}

We evaluate the performance of the proposed model, namely BDKN, on a set of standard UCI regression benchmarks \citep{ucirepo} and compare it against a diverse set of baseline methods, including Gaussian Processes (GP) \citep{williams2006gaussian}, Deep Ensembles (DE) \citep{lakshminarayanan2017simple}, Deep Kernel Learning (DKL) \citep{wilson2016deep}, Bayesian Last-Layer (BLL), Latent Derivative Bayesian Last-Layer (LD-BLL), and Mean-Field Variational Inference Bayesian Neural Networks (MFVI-BNN) \citep{watson2021latent}, and Variational Bayesian Last-Layer (VBLL) \citep{harrison2024variational}. The UCI datasets are widely used benchmarks \citep{lakshminarayanan2017simple,watson2021latent,harrison2024variational} for probabilistic nonlinear regression, covering a range of regression problems with varying input dimensionalities and sample sizes.

We report the root mean squared error (RMSE) to evaluate predictive accuracy and the negative log-likelihood (NLL) to assess uncertainty calibration. RMSE is defined as $\mathrm{RMSE} = \sqrt{\frac{1}{N}\sum_{i=1}^{N}(y_i - \hat{y}_i)^2}$, where $\hat{y}_i = \mathbb{E}[y_i \mid \x_i]$ denotes the predictive mean. NLL is computed as $\mathrm{NLL} = -\frac{1}{N}\sum_{i=1}^{N}\log p(y_i \mid \x_i)$, where $p(y_i \mid \x_i)$ denotes the predictive distribution of the model.

Experiments are conducted following \cite{lakshminarayanan2017simple}, using 20 repeated random training/test splits. For each split, 90\% of the data are used for training, and the remaining 10\% for testing. We report the mean and standard deviation of the evaluation metrics over the 20 splits; for the Protein dataset, which is substantially larger, 5 repeated splits are used, and for the Song dataset, a single split is used. More details about the datasets are provided in Appendix~\ref{subsub:data}. All methods are evaluated on the benchmark under two experimental settings: an aggressive optimization configuration and a more conservative configuration. 
In both settings, all neural-network-based approaches use two fully connected hidden layers with 50 neurons each and are trained with a batch size of 32. The only exception is the low-learning-rate setting on Sarcos, where BLL, LDBLL, and MFVI-BNN use three hidden layers with widths \(50\), \(200\), and \(200\), respectively. Our proposed model uses \(H=5\) feature maps (ensemble members). Additional details for each experimental setting and method, including activation functions, are provided in Appendix~\ref{subsub:settings}.

In the first setting, we use a learning rate of \(0.1\) and train all models for 40 epochs, following \cite{lakshminarayanan2017simple}, corresponding to a commonly used lightweight configuration in deep-ensemble studies and is widely adopted in practice. Results for this setting are reported in Table~\ref{tab:setting1_rmse_nll}.
% Detailed configurations are provided in the Appendix.
BDKN achieves competitive performance across datasets in both RMSE and NLL, and consistently outperforms DE. While DE exhibits large variance and unstable behavior on several datasets (e.g., Power, Sarcos), BDKN remains stable, demonstrating robustness to variability in optimization quality.
GP remains a strong baseline on several datasets. However, due to its cubic computational scaling and prohibitive memory requirements, we could not evaluate GP on large datasets such as Protein, Sarcos, and Song. Compared to other BLL-based methods (BLL, LDBLL) and variational baselines (MFVI, VBLL), BDKN generally provides a better balance between predictive accuracy and uncertainty calibration.

\begin{table}[t]
\centering
\scriptsize
\setlength{\tabcolsep}{4pt}
\renewcommand{\arraystretch}{1.1}

\caption{
\textbf{Experimental setting~1 (high learning rate): performance on standard regression benchmarks.}
\textbf{(a)} RMSE and \textbf{(b)} NLL reported as mean $\pm$ standard deviation across splits.
Lower values are better for both metrics. \textbf{Bold} indicates the best performance, and \underline{underlined} indicates the second-best performance. GP could not be evaluated on the Protein, Sarcos, and Song datasets, and LDBLL on the Song dataset, due to their high memory requirements, which exceeded the available GPU memory.
}
\begin{subtable}[t]{\textwidth}
\centering
\caption{RMSE (mean $\pm$ std).}
\label{tab:rmse_setting1_sub}
\resizebox{\textwidth}{!}{%
\begin{tabular}{lcccccccc}
\toprule
Dataset & BDKN & DE & LDBLL & GP & DKL & BLL & MFVI & VBLL \\
\midrule
Boston
& \underline{3.03 $\pm$ 0.78}
& 3.60 $\pm$ 1.76
& 7.80 $\pm$ 3.90
& \textbf{2.83 $\pm$ 0.76}
& 3.45 $\pm$ 0.81
& 4.04 $\pm$ 1.67
& 5.45 $\pm$ 0.99
& 3.68 $\pm$ 1.05 \\
Concrete
& \underline{5.90 $\pm$ 0.48}
& 7.18 $\pm$ 3.46
& 13.71 $\pm$ 6.76
& \textbf{5.64 $\pm$ 0.66}
& 7.18 $\pm$ 0.67
& 8.84 $\pm$ 3.20
& 12.97 $\pm$ 2.41
& 6.62 $\pm$ 1.19 \\
Energy
& \underline{1.38 $\pm$ 0.33}
& 2.54 $\pm$ 1.11
& 6.67 $\pm$ 6.10
& \textbf{0.65 $\pm$ 0.06}
& 2.65 $\pm$ 0.42
& 5.77 $\pm$ 4.82
& 5.64 $\pm$ 1.78
& 1.64 $\pm$ 0.37 \\
Kin8nm
& 0.15 $\pm$ 0.04
& 0.22 $\pm$ 0.04
& 0.72 $\pm$ 1.01
& \textbf{0.07 $\pm$ 0.00}
& \underline{0.10 $\pm$ 0.02}
& \underline{0.10 $\pm$ 0.00}
& 0.23 $\pm$ 0.03
& 0.19 $\pm$ 0.03 \\
Naval
& \underline{0.01 $\pm$ 0.00}
& 3.17 $\pm$ 7.88
& 0.03 $\pm$ 0.06
& \textbf{0.00 $\pm$ 0.00}
& \underline{0.01 $\pm$ 0.00}
& \textbf{0.00 $\pm$ 0.01}
& \underline{0.01 $\pm$ 0.00}
& 0.33 $\pm$ 0.33 \\
Power
& 4.79 $\pm$ 1.02
& 3424.99 $\pm$ 6918.44
& 14.93 $\pm$ 10.67
& \textbf{3.95 $\pm$ 0.16}
& \underline{4.44 $\pm$ 0.19}
& 6.33 $\pm$ 3.26
& 9.46 $\pm$ 1.86
& 8.38 $\pm$ 3.16 \\
Protein
& \underline{4.94 $\pm$ 0.09}
& 5.31 $\pm$ 0.25
& 12.85 $\pm$ 7.03
& N/A
& 5.03 $\pm$ 0.10
& \textbf{4.42 $\pm$ 0.16}
& 5.58 $\pm$ 0.16
& \underline{4.94 $\pm$ 0.09} \\
Sarcos
& \underline{3.94 $\pm$ 0.18}
& 53.62 $\pm$ 157.00
& 12.11 $\pm$ 1.36
& N/A
& 5.61 $\pm$ 1.06
& \textbf{3.68 $\pm$ 0.06}
& 15.44 $\pm$ 3.32
& 5.47 $\pm$ 1.24 \\
Song
& 10.85 $\pm$ 0.00
& 10.86 $\pm$ 0.00
& N/A
& N/A
& \underline{8.98 $\pm$ 0.00}
& \textbf{8.91 $\pm$ 0.00}
& 11.60 $\pm$ 0.00
& 10.85 $\pm$ 0.00 \\
Wine
& \underline{0.64 $\pm$ 0.04}
& 1.97 $\pm$ 3.03
& 0.78 $\pm$ 0.09
& \textbf{0.56 $\pm$ 0.04}
& 0.65 $\pm$ 0.04
& 0.80 $\pm$ 0.05
& 0.73 $\pm$ 0.06
& 0.71 $\pm$ 0.04 \\
Yacht
& \textbf{0.82 $\pm$ 0.38}
& 1.68 $\pm$ 1.06
& 4.16 $\pm$ 4.20
& 1.88 $\pm$ 0.43
& \underline{1.22 $\pm$ 0.64}
& 4.72 $\pm$ 1.06
& 9.42 $\pm$ 2.99
& 1.70 $\pm$ 0.57 \\
\bottomrule
\end{tabular}
}
\end{subtable}

\vspace{0.6em}

\begin{subtable}[t]{\textwidth}
\centering
\caption{NLL (mean $\pm$ std).}
\label{tab:nll_setting1_sub}
\resizebox{\textwidth}{!}{%
\begin{tabular}{lcccccccc}
\toprule
Dataset & BDKN & DE & LDBLL & GP & DKL & BLL & MFVI & VBLL \\
\midrule
Boston
& \underline{2.61 $\pm$ 0.42}
& 3.03 $\pm$ 0.79
& 4.73 $\pm$ 0.31
& \textbf{2.42 $\pm$ 0.28}
& 4.45 $\pm$ 1.60
& 3.32 $\pm$ 0.14
& 3.24 $\pm$ 0.16
& 2.83 $\pm$ 0.49 \\
Concrete
& \textbf{3.22 $\pm$ 0.10}
& 3.67 $\pm$ 0.67
& 4.81 $\pm$ 0.25
& 3.53 $\pm$ 0.04
& 13.16 $\pm$ 2.07
& 3.94 $\pm$ 0.15
& 4.03 $\pm$ 0.20
& \underline{3.37 $\pm$ 0.28} \\
Energy
& \underline{1.73 $\pm$ 0.24}
& 2.79 $\pm$ 0.60
& 3.39 $\pm$ 1.31
& \textbf{1.40 $\pm$ 0.02}
& 3.42 $\pm$ 0.66
& 3.44 $\pm$ 0.23
& 3.27 $\pm$ 0.20
& 1.96 $\pm$ 0.30 \\
Kin8nm
& \underline{-0.48 $\pm$ 0.22}
& -0.02 $\pm$ 0.30
& 1.29 $\pm$ 1.12
& \textbf{-0.68 $\pm$ 0.00}
& 393.44 $\pm$ 953.78
& -0.34 $\pm$ 0.00
& 0.01 $\pm$ 0.15
& -0.21 $\pm$ 0.24 \\
Naval
& -2.80 $\pm$ 0.01
& -0.07 $\pm$ 2.83
& -2.05 $\pm$ 0.88
& \textbf{-5.11 $\pm$ 0.01}
& -1.14 $\pm$ 0.00
& \underline{-3.20 $\pm$ 0.22}
& -3.03 $\pm$ 0.18
& -0.51 $\pm$ 1.50 \\
Power
& \underline{2.97 $\pm$ 0.17}
& 6.89 $\pm$ 2.06
& 5.46 $\pm$ 2.89
& \textbf{2.83 $\pm$ 0.03}
& 6.91 $\pm$ 0.62
& 3.85 $\pm$ 0.11
& 3.88 $\pm$ 0.22
& 3.97 $\pm$ 1.57 \\
Protein
& \underline{3.00 $\pm$ 0.02}
& 3.12 $\pm$ 0.11
& 4.21 $\pm$ 0.08
& N/A
& 6.11 $\pm$ 0.15
& \textbf{2.99 $\pm$ 0.02}
& 3.17 $\pm$ 0.04
& 3.03 $\pm$ 0.03 \\
Sarcos
& \textbf{2.80 $\pm$ 0.04}
& 4.02 $\pm$ 1.52
& 5.23 $\pm$ 0.44
& N/A
& 9.71 $\pm$ 2.74
& 3.95 $\pm$ 0.00
& 4.16 $\pm$ 0.18
& \underline{3.11 $\pm$ 0.23} \\
Song
& \textbf{3.49 $\pm$ 0.00}
& 3.81 $\pm$ 0.00
& N/A
& N/A
& \underline{3.61 $\pm$ 0.00}
& 3.65 $\pm$ 0.00
& 4.05 $\pm$ 0.00
& 3.81 $\pm$ 0.00 \\
Wine
& \underline{0.98 $\pm$ 0.06}
& 1.59 $\pm$ 1.08
& 2.07 $\pm$ 0.23
& \textbf{0.86 $\pm$ 0.05}
& \underline{0.98 $\pm$ 0.07}
& 1.19 $\pm$ 0.06
& 1.12 $\pm$ 0.10
& 1.08 $\pm$ 0.06 \\
Yacht
& \textbf{1.23 $\pm$ 0.55}
& 2.10 $\pm$ 0.73
& 2.79 $\pm$ 0.47
& 2.07 $\pm$ 0.08
& \underline{1.74 $\pm$ 0.71}
& 3.77 $\pm$ 0.09
& 3.75 $\pm$ 0.25
& 1.89 $\pm$ 0.45 \\
\bottomrule
\end{tabular}
}
\end{subtable}

\label{tab:setting1_rmse_nll}
\end{table}

% , a batch size of 32, and a maximum of 500 training epochs. 
% We choose this lightweight configuration as it converges rapidly in practice while yielding strong empirical performance across datasets. 
% In this first experimental setting, we employ comparable network architectures and learning rates for the other NN-based baselines. \todo{more info in appendix}

To also consider scenarios in which more conservative optimization is used, we introduce a second experimental setting with lower learning rates. In this setting, we follow the benchmark configurations proposed by \citep{watson2021latent}, which use dataset-specific learning rates and training schedules. This setting allows us to demonstrate the benefits of our method even when optimization is less aggressive, and to assess robustness to changes in optimization hyperparameters. The exact configurations, including learning rates, numbers of epochs, and training procedures for each dataset, are provided in Appendix~\ref{subsub:settings}. 

The empirical performance of all methods under this second experimental setting is summarized in Table~\ref{tab:setting2_rmse_nll}, where methods exhibit increased stability under this more conservative optimization regime. BDKN remains competitive in both RMSE and NLL and continues to provide well-calibrated uncertainty across datasets. DE performs strongly in this setting, with reduced variations in optimization quality, while GP again achieves strong performance on several benchmarks. In contrast, BLL-based and variational methods exhibit dataset-dependent behavior. The results obtained for LDBLL and VBLL are consistent with those reported in \cite{harrison2024variational}. MFVI also shows strong and stable performance as well, but at a substantially higher computational cost. Runtime comparisons for both experimental settings are reported in Appendix~\ref{subsub:result}.

\begin{table}[t]
\centering
\scriptsize
\setlength{\tabcolsep}{4pt}
\renewcommand{\arraystretch}{1.1}

\caption{
\textbf{Experimental setting~2 (low learning rate): performance on standard regression benchmarks.}
\textbf{(a)} RMSE and \textbf{(b)} NLL reported as mean $\pm$ standard deviation across splits.
Lower values indicate better performance. \textbf{Bold} indicates the best performance, and \underline{underlined} indicates the second-best performance. GP could not be evaluated on the Protein, Sarcos, and Song datasets, and LDBLL on the Song dataset, due to their high memory requirements, which exceeded the available GPU memory.
}

% -------------------- RMSE --------------------
\begin{subtable}[t]{\textwidth}
\centering
\caption{RMSE (mean $\pm$ std).}
\label{tab:rmse_setting2_sub}
\resizebox{\textwidth}{!}{%
\begin{tabular}{lcccccccc}
\toprule
Dataset & BDKN & DE & LDBLL & GP & DKL & BLL & MFVI & VBLL \\
\midrule
Boston
& 3.01 $\pm$ 0.78
& 2.99 $\pm$ 0.85
& 4.41 $\pm$ 1.20
& \textbf{2.83 $\pm$ 0.73}
& 4.11 $\pm$ 1.18
& 3.20 $\pm$ 0.73
& \underline{2.89 $\pm$ 0.68}
& 2.92 $\pm$ 0.77 \\
Concrete
& 5.22 $\pm$ 0.58
& 5.23 $\pm$ 0.59
& 7.03 $\pm$ 0.58
& 5.63 $\pm$ 0.62
& 5.83 $\pm$ 0.93
& 5.24 $\pm$ 0.69
& \textbf{4.55 $\pm$ 0.60}
& \underline{5.11 $\pm$ 0.62} \\
Energy
& 0.76 $\pm$ 0.21
& 1.42 $\pm$ 0.29
& 0.91 $\pm$ 0.71
& \underline{0.47 $\pm$ 0.06}
& 1.02 $\pm$ 0.50
& \underline{0.47 $\pm$ 0.07}
& \textbf{0.43 $\pm$ 0.07}
& 0.48 $\pm$ 0.06 \\
Kin8nm
& \textbf{0.07 $\pm$ 0.00}
& \textbf{0.07 $\pm$ 0.00}
& 0.10 $\pm$ 0.03
& \textbf{0.07 $\pm$ 0.00}
& \underline{0.09 $\pm$ 0.01}
& 0.10 $\pm$ 0.00
& \textbf{0.07 $\pm$ 0.00}
& \textbf{0.07 $\pm$ 0.00} \\
Naval
& \textbf{0.00 $\pm$ 0.00}
& \textbf{0.00 $\pm$ 0.00}
& \textbf{0.00 $\pm$ 0.00}
& \textbf{0.00 $\pm$ 0.00}
& \underline{0.01 $\pm$ 0.00}
& \textbf{0.00 $\pm$ 0.00}
& \textbf{0.00 $\pm$ 0.00}
& \textbf{0.00 $\pm$ 0.00} \\
Power
& 3.93 $\pm$ 0.16
& 3.98 $\pm$ 0.16
& \underline{3.87 $\pm$ 0.18}
& \textbf{3.72 $\pm$ 0.17}
& 4.36 $\pm$ 0.51
& 4.05 $\pm$ 0.17
& \underline{3.87 $\pm$ 0.15}
& 4.22 $\pm$ 0.20 \\
Protein
& 4.05 $\pm$ 0.02
& 4.06 $\pm$ 0.03
& 4.17 $\pm$ 0.14
& N/A
& 4.22 $\pm$ 0.06
& 4.09 $\pm$ 0.05
& \textbf{3.87 $\pm$ 0.01}
& \underline{3.95 $\pm$ 0.04} \\
Sarcos
& 2.13 $\pm$ 0.03
& 2.14 $\pm$ 0.03
& 9.25 $\pm$ 1.13
& N/A
& 3.46 $\pm$ 0.90
& \textbf{1.61 $\pm$ 0.06}
& \underline{2.09 $\pm$ 0.07}
& 2.21 $\pm$ 0.05 \\
Song
& \underline{8.71 $\pm$ 0.00}
& \textbf{8.69 $\pm$ 0.00}
& N/A
& N/A
& 9.08 $\pm$ 0.00
& 9.05 $\pm$ 0.00
& 9.61 $\pm$ 0.00
& 9.14 $\pm$ 0.00 \\
Wine
& 0.65 $\pm$ 0.06
& \underline{0.63 $\pm$ 0.05}
& 0.65 $\pm$ 0.03
& \textbf{0.56 $\pm$ 0.04}
& 0.76 $\pm$ 0.06
& 0.64 $\pm$ 0.04
& \underline{0.63 $\pm$ 0.04}
& 0.72 $\pm$ 0.06 \\
Yacht
& 0.85 $\pm$ 0.38
& 1.17 $\pm$ 0.44
& 3.25 $\pm$ 5.68
& \textbf{0.40 $\pm$ 0.13}
& 2.12 $\pm$ 1.89
& 2.36 $\pm$ 0.52
& 0.86 $\pm$ 0.42
& \underline{0.63 $\pm$ 0.29} \\
\bottomrule
\end{tabular}
}
\end{subtable}

\vspace{0.6em}

% -------------------- NLL --------------------
\begin{subtable}[t]{\textwidth}
\centering
\caption{NLL (mean $\pm$ std).}
\label{tab:nll_setting2_sub}
\resizebox{\textwidth}{!}{%
\begin{tabular}{lcccccccc}
\toprule
Dataset & BDKN & DE & LDBLL & GP & DKL & BLL & MFVI & VBLL \\
\midrule
Boston
& 2.81 $\pm$ 0.62
& \underline{2.52 $\pm$ 0.37}
& 3.30 $\pm$ 0.05
& \textbf{2.41 $\pm$ 0.28}
& 13.05 $\pm$ 7.29
& 3.21 $\pm$ 0.03
& 3.27 $\pm$ 0.95
& 3.25 $\pm$ 1.16 \\
Concrete
& \underline{3.23 $\pm$ 0.22}
& \textbf{3.01 $\pm$ 0.18}
& 3.83 $\pm$ 0.02
& 3.42 $\pm$ 0.04
& 23.59 $\pm$ 7.38
& 3.79 $\pm$ 0.01
& 3.80 $\pm$ 0.69
& 3.24 $\pm$ 0.26 \\
Energy
& 1.17 $\pm$ 0.30
& 1.71 $\pm$ 1.74
& 1.56 $\pm$ 0.49
& \textbf{0.66 $\pm$ 0.16}
& 1.62 $\pm$ 1.23
& 3.23 $\pm$ 0.00
& 3.25 $\pm$ 1.32
& \underline{0.88 $\pm$ 0.27} \\
Kin8nm
& \underline{-1.18 $\pm$ 0.07}
& \textbf{-1.35 $\pm$ 0.03}
& -0.68 $\pm$ 0.21
& -0.58 $\pm$ 0.05
& 0.55 $\pm$ 0.00
& -0.34 $\pm$ 0.01
& -1.13 $\pm$ 0.04
& -0.97 $\pm$ 0.07 \\
Naval
& -4.76 $\pm$ 0.03
& \underline{-5.33 $\pm$ 0.18}
& -4.59 $\pm$ 0.76
& -4.70 $\pm$ 0.01
& 0.54 $\pm$ 0.00
& -3.30 $\pm$ 0.00
& \textbf{-8.10 $\pm$ 0.19}
& -3.70 $\pm$ 0.01 \\
Power
& \underline{2.79 $\pm$ 0.05}
& \underline{2.79 $\pm$ 0.04}
& \textbf{2.78 $\pm$ 0.05}
& 3.19 $\pm$ 0.01
& 12.11 $\pm$ 3.06
& 3.78 $\pm$ 0.00
& \textbf{2.78 $\pm$ 0.04}
& 2.86 $\pm$ 0.05 \\
Protein
& \underline{2.80 $\pm$ 0.01}
& \textbf{2.74 $\pm$ 0.23}
& 2.86 $\pm$ 0.03
& N/A
& 10.68 $\pm$ 0.25
& 2.95 $\pm$ 0.01
& 2.81 $\pm$ 0.01
& \underline{2.80 $\pm$ 0.01} \\
Sarcos
& 2.19 $\pm$ 0.02
& \textbf{1.83 $\pm$ 0.01}
& 4.14 $\pm$ 0.43
& N/A
& 9.04 $\pm$ 5.00
& 3.94 $\pm$ 0.00
& \underline{2.17 $\pm$ 0.03}
& 2.21 $\pm$ 0.02 \\
Song
& \textbf{3.30 $\pm$ 0.00}
& \underline{3.33 $\pm$ 0.00}
& N/A
& N/A
& 3.63 $\pm$ 0.00
& 3.66 $\pm$ 0.00
& 3.99 $\pm$ 0.00
& 3.65 $\pm$ 0.00 \\
Wine
& 1.64 $\pm$ 0.36
& 1.11 $\pm$ 0.24
& 1.03 $\pm$ 0.03
& \textbf{0.86 $\pm$ 0.06}
& 1.16 $\pm$ 0.11
& 1.02 $\pm$ 0.03
& \underline{0.94 $\pm$ 0.06}
& 3.32 $\pm$ 0.67 \\
Yacht
& 1.48 $\pm$ 0.84
& \textbf{-0.20 $\pm$ 0.26}
& 1.79 $\pm$ 0.39
& \underline{0.17 $\pm$ 0.12}
& 6.12 $\pm$ 13.77
& 3.66 $\pm$ 0.01
& 1.41 $\pm$ 1.20
& 2.22 $\pm$ 2.63 \\
\bottomrule
\end{tabular}
}
\end{subtable}

\label{tab:setting2_rmse_nll}
\end{table}

\begin{figure}[h]
    \centering
    \begin{subfigure}[t]{0.435\textwidth}
        \includegraphics[width=\linewidth]{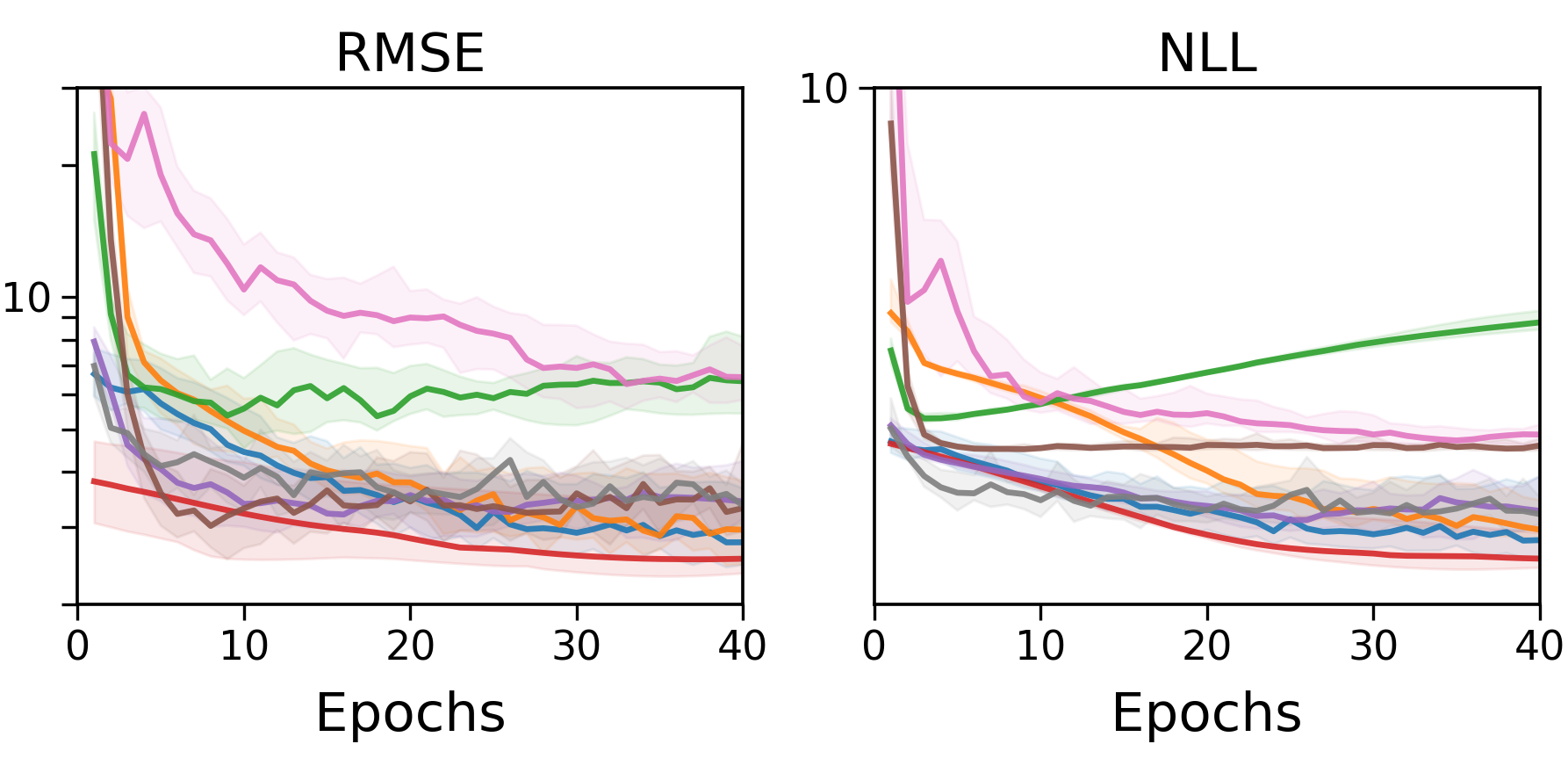}
        
        \caption{high learning rate}
        \label{fig:boston_setting1_test_epochs}
    \end{subfigure}
    \hfill
    \begin{subfigure}[t]{0.435\textwidth}
        \includegraphics[width=\linewidth]{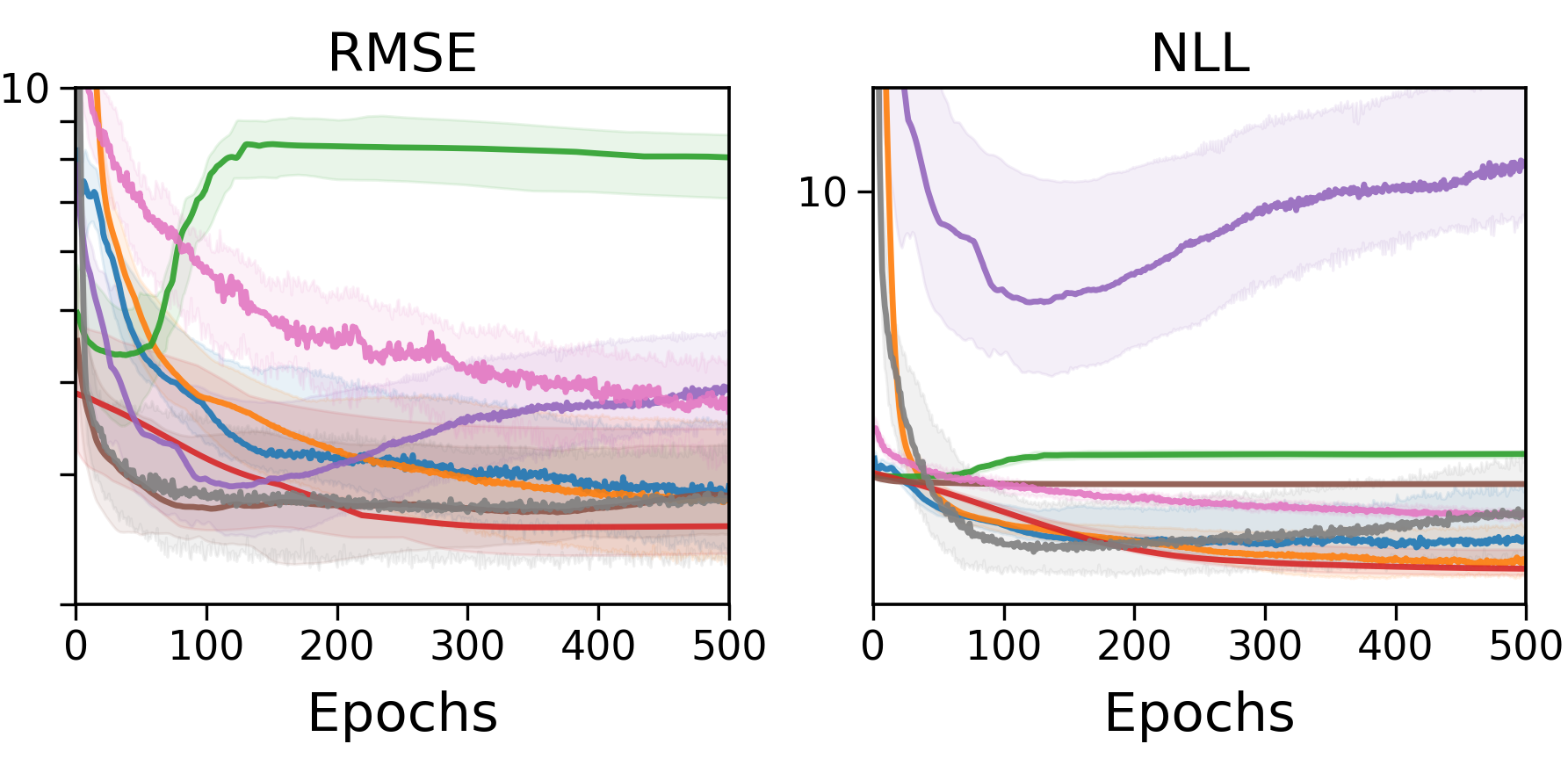}
        \caption{low learning rate}
        \label{fig:boston_setting2_test_epochs}
    \end{subfigure}
    \hfill    
    \begin{subfigure}[t]{0.115\textwidth}
          \raisebox{0.2\height}{\includegraphics[width=\linewidth]{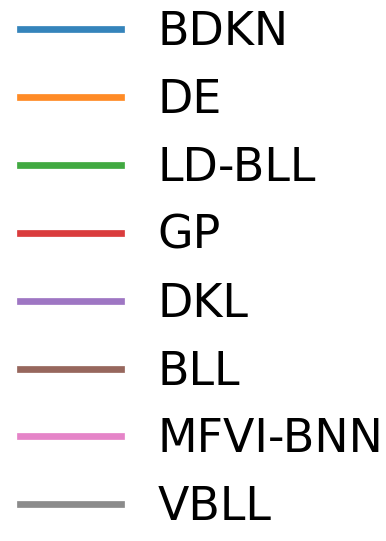}}
        % \caption{low learning rate}
        % \label{fig:boston_setting2_test_epochs}
    \end{subfigure}
    \caption{
    \textbf{Boston dataset - Benchmarks test data performance over training epochs under two experimental settings.}
    \textbf{(a)} Setting~1 (high learning rate).
    \textbf{(b)} Setting~2 (low learning rate).
    We show quartile plots of \textbf{test RMSE} (left) and \textbf{test NLL} (right) as functions of training epochs.
    Solid lines denote the median across 20 splits, while shaded regions indicate the interquartile range (25th--75th percentiles).
    Lower values are better for both RMSE and NLL.
    }
    \label{fig:boston_test_epochs_settings}
\end{figure}

Comparing the two experimental settings, BDKN exhibits robust behavior across both high and low learning-rate regimes. Figure~\ref{fig:boston_test_epochs_settings} compares test RMSE and NLL over training epochs on the Boston dataset under the two settings.
In the high learning-rate setting, all models exhibit stable training dynamics and rapid convergence, except for LDBLL, which degrades after a few epochs (Figure~\ref{fig:boston_setting1_test_epochs}). Moreover, MFVI continues to converge slowly despite the aggressive learning rate.
Under the low learning-rate setting, optimization is more stable across methods, with smoother convergence and reduced variance. Figure~\ref{fig:boston_setting2_test_epochs} shows the test-data evaluation over the first 500 training epochs for all methods. GP demonstrates good performance, although with slower convergence. BLL exhibits limited improvement, converging quickly but to higher RMSE and NLL values. LDBLL initially degrades after the first few epochs and then begins to converge very slowly. DKL also shows degradation after a few epochs. In contrast, BDKN and DE maintain fast and stable convergence throughout training. Across both settings, BDKN maintains stable training dynamics and consistent performance in RMSE and NLL, further supporting its robustness.

% \begin{figure}[h]
%     \centering
%     \begin{subfigure}[t]{0.435\textwidth}
%         \includegraphics[width=\linewidth]{fig/trimed_fig2a_N_500.png}
        
%         \caption{high learning rate}
%         \label{fig:boston_setting1_test_epochs}
%     \end{subfigure}
%     \hfill
%     \begin{subfigure}[t]{0.435\textwidth}
%         \includegraphics[width=\linewidth]{fig/trimed_fig2b_N.png}
%         \caption{low learning rate}
%         \label{fig:boston_setting2_test_epochs}
%     \end{subfigure}
%     \hfill    
%     \begin{subfigure}[t]{0.115\textwidth}
%         % \includegraphics[width=\linewidth]{fig/fig2c_legend.png}
%           \raisebox{0.2\height}{\includegraphics[width=\linewidth]{fig/fig2c_legend_N.png}}
%         % \caption{low learning rate}
%         % \label{fig:boston_setting2_test_epochs}
%     \end{subfigure}
%     \caption{
%     \textbf{Boston dataset - Benchmarks test data performance over training epochs under two experimental settings.}
%     \textbf{(a)} Setting~1 (high learning rate).
%     \textbf{(b)} Setting~2 (low learning rate).
%     We show quartile plots of \textbf{test RMSE} (left) and \textbf{test NLL} (right) as functions of training epochs.
%     Solid lines denote the median across 20 splits, while shaded regions indicate the interquartile range (25th--75th percentiles).
%     Lower values are better for both RMSE and NLL.
%     }
%     \label{fig:boston_test_epochs_settings}
% \end{figure}

\begin{figure}[t]
    \centering
    \begin{subfigure}[b]{0.49\textwidth}
        \includegraphics[width=\linewidth]{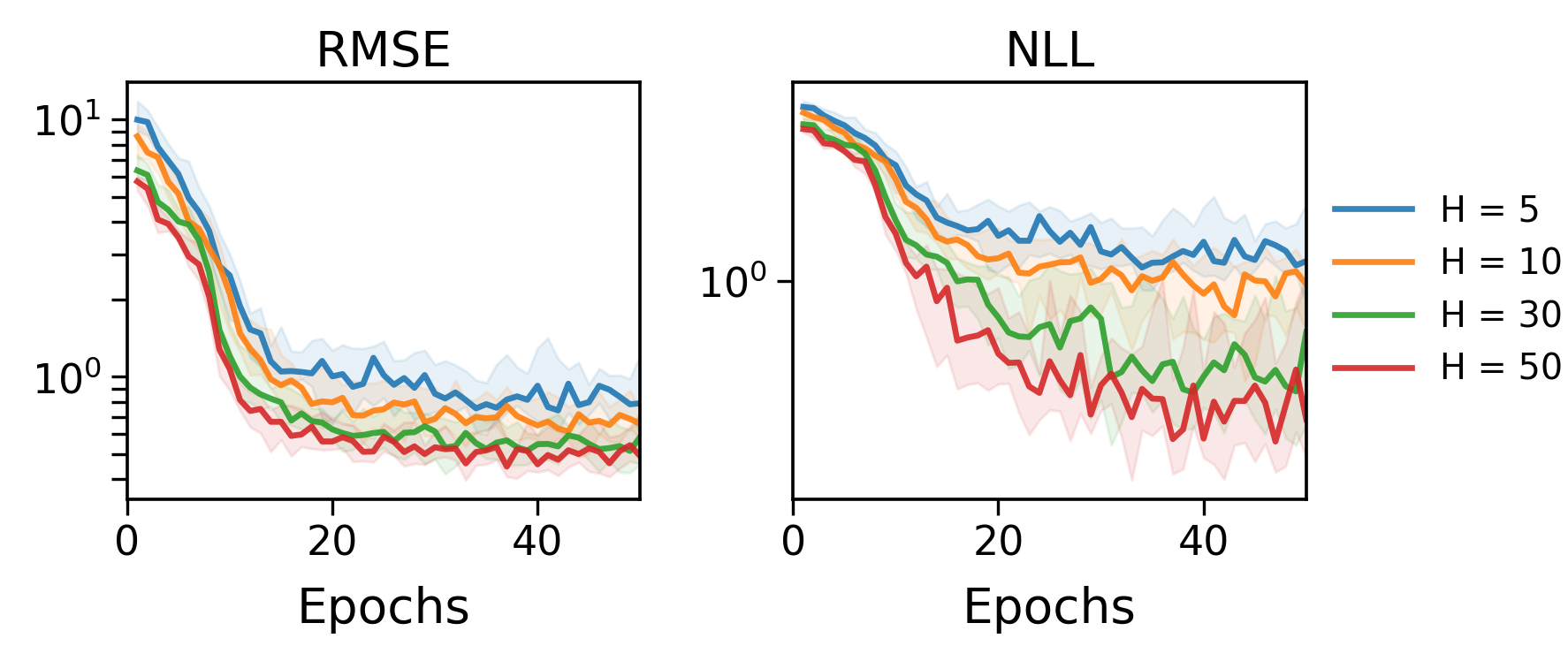}
        \caption{}
        \label{fig:yacht_setting1_epochs}
    \end{subfigure}
    \hfill
    \begin{subfigure}[b]{0.49\textwidth}
    \includegraphics[width=\linewidth, height=0.42\linewidth, keepaspectratio]{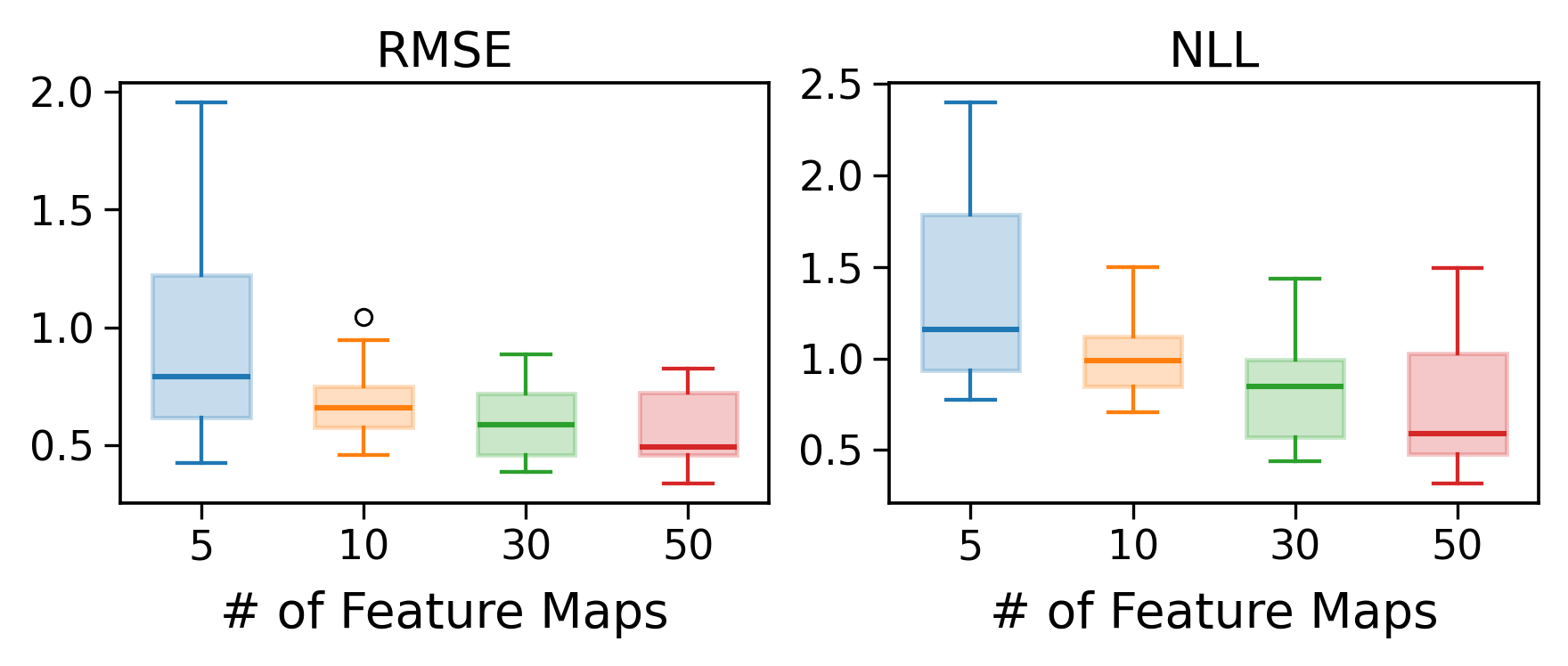}
        \caption{}
        \label{fig:yacht_setting1_final}
    \end{subfigure}

    \caption{
    \textbf{Yacht (Setting~1): effect of BDKN ensemble size on the test data performance.}
    \textbf{(a)} Quartile plots of \textbf{test RMSE} (left) and \textbf{test NLL} (right) as functions of training epochs for ensembles with 5, 10, 30, and 50 feature maps.
    Solid lines denote the median across splits and shaded regions indicate the interquartile range (25th--75th percentiles).
    \textbf{(b)} Boxplots of \textbf{test RMSE} (left) and \textbf{test NLL} (right) across splits at the final training epoch.
    Boxes show the interquartile range with the median marked; whiskers denote the full range excluding outliers.
    Lower values are better for both RMSE and NLL.
    }
    \label{fig:yacht_setting1_combined}
\end{figure}

A key advantage of BDKN is that its feature maps can be trained fully in parallel as ensemble members. Although we use five ensemble members following \cite{lakshminarayanan2017simple} for simplicity and scalability, the ensemble size can be increased with parallel computing without substantially increasing training time.
Figure~\ref{fig:yacht_setting1_combined} illustrates the effect of ensemble size on performance for the Yacht dataset under Setting~1 with 50 training epochs. 
As the number of ensemble members increases, both RMSE and NLL improve consistently, and larger ensembles exhibit faster convergence across splits.
This behavior is evident throughout training (Figure~\ref{fig:yacht_setting1_epochs}) and at the final epoch, where the median RMSE and NLL decrease monotonically with ensemble size (Figure~\ref{fig:yacht_setting1_final}). At the final epoch, Figure~\ref{fig:yacht_setting1_final} shows a clear decrease in the median RMSE and NLL as the ensemble size increases. These results indicate that increasing the ensemble size yields a richer and more expressive feature space, improving predictive accuracy and uncertainty calibration without a significant increase in training time when sufficient computational resources are available.

\section{Related work}

% \subsection{Gaussian Process}
\noindent\textbf{Gaussian process.}
Gaussian Processes (GPs) provide a nonparametric Bayesian framework for regression by placing a prior directly over functions \cite{williams2006gaussian}. However, their practical use is often limited by computational complexity and dependence on kernel choice \citep{quinonero2005unifying,wilson2013gaussian}, motivating a large body of approximation methods such as sparse and inducing-point approaches \citep{snelson2005sparse,titsias2009variational}. Beyond scalability, numerous extensions have been proposed to improve expressiveness, including neural-network-based kernels and deep GP variants \citep{cho2009kernel,calandra2016manifold,wilson2016deep,bui2016deep}. Related theoretical work has also established connections between infinitely wide neural networks and GPs, characterizing the function-space priors induced by neural architectures \citep{lee2017deep,rudner2018connection,khan2019approximate}.

\noindent\textbf{Bayesian neural networks.}
BNNs place prior distributions over network parameters and perform posterior inference to capture predictive uncertainty \citep{neal2012bayesian}. While fully Bayesian inference via Monte Carlo sampling methods \cite{andrieu2003introduction, hoffman2014no, chen2014stochastic} provides a principled treatment of uncertainty, such approaches are typically computationally expensive and difficult to scale to modern architectures \citep{welling2011bayesian}. As a result, much of the recent work on BNNs has focused on scalable approximate inference techniques, most notably variational inference methods, including mean-field approximations, which trade posterior fidelity for computational tractability \citep{shridhar2019comprehensive, coker2022wide, sun2019functional, pmlr-v286-cinquin25a}.

\noindent\textbf{Bayesian last layer networks.}
BLLs restrict Bayesian inference to a linear head on top of learned deterministic features, offering a practical trade-off between scalability and uncertainty estimation. 
Building on this formulation, several extensions refine training and uncertainty behavior.  
In particular, an efficient reformulation of BLL enables backpropagation-based marginal likelihood training and improves extrapolative uncertainty quantification \cite{fiedler2023improved}. To address overconfident predictions and improve uncertainty estimation outside the training distribution, LD-BLL networks introduce functional priors over network derivatives with respect to the inputs to encourage more expressive and calibrated uncertainty \cite{watson2021latent}. More recently, VBLL has been proposed to further improve scalability and uncertainty estimation through a sampling-free, single-pass training objective \cite{harrison2024variational}.

\noindent\textbf{Deep ensembles.}
DE combines independently trained neural networks, typically via uniform averaging, and is widely used in practice, including reinforcement learning \citep{osband2018randomized, an2020deep,wu2020deep, lee2021sunrise, abe2022deep, seligmann2023beyond}. 
% Despite their promising performance, their averaging aggregation introduces two practical limitations \citep{fort2019deep, pearce2020uncertainty, d2021repulsive}. First, uniform averaging is non-diagnostic. While averaging can improve accuracy and NLL, it does not provide a probabilistic notion of which members are supported by the data, nor does it enable attribution of predictive behavior to specific predictors. 
% Second, DE can be brittle under optimization variability. In practice, independently trained members may exhibit heterogeneous quality due to sensitivity to initialization, stochastic optimization, or aggressive training settings, and a uniformly averaged ensemble can inherit overconfident failures that disproportionately degrade NLL. 
Several studies have provided theoretical and analytical perspectives on standard deep ensembles and their uncertainty behavior \citep{fort2019deep, pearce2020uncertainty, d2021repulsive}.
Prior work has shown that promoting ensemble diversity can improve uncertainty estimation in DE \citep{reeve2018diversity,nam2021diversity, rame2021dice}. For instance, ensemble diversity can be encouraged by augmenting the loss function with an auxiliary regularization term that maximizes ensemble predictive variance on inputs sampled uniformly from the input space, improving uncertainty behavior away from the training data \citep{jain2020maximizing}. Diversity can also be induced structurally through architectural heterogeneity across ensemble members, such as by randomizing activation functions across networks \citep{zaidi2021neural, stoyanova2023toward}.

% In contrast, our Bayesian aggregation places an explicit posterior over member contributions, allowing the posterior mass to be interpreted as evidence for or against particular ensemble members and enabling diagnostic analyses.
% Additionally, our approach is robust against quality variations by automatically downweighting members that are weakly supported by the observations, thereby improving robustness without manual member selection or additional tuning heuristics.

\section{Discussion}
\label{sec:discussion}

We introduced BDKN, a scalable framework that bridges the gap between the empirical success of DE and the statistical rigor of GPs. By interpreting independently trained neural networks as induced basis functions, BDKN constructs a finite-rank deep kernel model that enables exact Bayesian inference without the cubic computational costs associated with traditional GPs. Since the ensemble members in BDKN are trained independently, the feature maps can be learned in parallel. This allows the number of feature maps to be scaled to increase model expressivity without a significant increase in training time, which we empirically show leads to improved performance. Our approach offers a clear advantage over standard weight-space BNNs and variational approximations: it yields closed-form posterior predictive distributions together with an interpretable posterior over ensemble members. BDKN also addresses a key limitation of DE, namely their non-diagnostic uniform averaging, by replacing it with a Bayesian aggregation layer that automatically downweights ensemble members weakly supported by the data.
This aggregation mechanism improves robustness by reducing the impact of poorly optimized or low-quality ensemble members, which can otherwise degrade performance under aggressive training regimes.
While baselines such as DE and LD-BLL show sensitivity to learning-rate selection and can become unstable in high learning-rate regimes, BDKN consistently maintains competitive predictive accuracy and well-calibrated uncertainty across both conservative and aggressive training configurations.

This study focuses on regression; extending the BDKN framework to classification tasks is a natural next step. A key open challenge is maintaining analytic tractability of the posterior predictive distribution under non-Gaussian likelihoods while retaining the robustness and interpretability observed in the regression setting. Further improvements in uncertainty calibration may be achieved by enhancing variability in the learned feature maps or by incorporating principles from latent-derivative modeling. Finally, extending BDKN to other domains, such as deep reinforcement learning and physics-informed machine learning, represents a promising direction for future work.

\bibliography{example_paper,ref}

@book{williams2006gaussian,
  title={Gaussian processes for machine learning},
  author={Williams, Christopher KI and Rasmussen, Carl Edward},
  volume={2},
  number={3},
  year={2006},
  publisher={MIT press Cambridge, MA}
}

@inproceedings{wilson2016deep,
  title={Deep kernel learning},
  author={Wilson, Andrew Gordon and Hu, Zhiting and Salakhutdinov, Ruslan and Xing, Eric P},
  booktitle={Artificial intelligence and statistics},
  pages={370--378},
  year={2016},
  organization={PMLR}
}

@article{filos2019systematic,
  title={A systematic comparison of bayesian deep learning robustness in diabetic retinopathy tasks},
  author={Filos, Angelos and Farquhar, Sebastian and Gomez, Aidan N and Rudner, Tim GJ and Kenton, Zachary and Smith, Lewis and Alizadeh, Milad and De Kroon, Arnoud and Gal, Yarin},
  journal={arXiv preprint arXiv:1912.10481},
  year={2019}
}

@inproceedings{guo2017calibration,
  title={On calibration of modern neural networks},
  author={Guo, Chuan and Pleiss, Geoff and Sun, Yu and Weinberger, Kilian Q},
  booktitle={International conference on machine learning},
  pages={1321--1330},
  year={2017},
  organization={PMLR}
}

@article{lakshminarayanan2017simple,
  title={Simple and scalable predictive uncertainty estimation using deep ensembles},
  author={Lakshminarayanan, Balaji and Pritzel, Alexander and Blundell, Charles},
  journal={Advances in neural information processing systems},
  volume={30},
  year={2017}
}

@inproceedings{watson2021latent,
  title={Latent derivative Bayesian last layer networks},
  author={Watson, Joe and Lin, Jihao Andreas and Klink, Pascal and Pajarinen, Joni and Peters, Jan},
  booktitle={International Conference on Artificial Intelligence and Statistics},
  pages={1198--1206},
  year={2021},
  organization={PMLR}
}

@book{neal2012bayesian,
  title={Bayesian learning for neural networks},
  author={Neal, Radford M},
  volume={118},
  year={2012},
  publisher={Springer Science \& Business Media}
}

@article{wilson2020bayesian,
  title={Bayesian deep learning and a probabilistic perspective of generalization},
  author={Wilson, Andrew G and Izmailov, Pavel},
  journal={Advances in neural information processing systems},
  volume={33},
  pages={4697--4708},
  year={2020}
}

@article{foong2020expressiveness,
  title={On the expressiveness of approximate inference in bayesian neural networks},
  author={Foong, Andrew and Burt, David and Li, Yingzhen and Turner, Richard},
  journal={Advances in Neural Information Processing Systems},
  volume={33},
  pages={15897--15908},
  year={2020}
}

@article{graves2011practical,
  title={Practical variational inference for neural networks},
  author={Graves, Alex},
  journal={Advances in neural information processing systems},
  volume={24},
  year={2011}
}

@article{wenzel2020good,
  title={How good is the bayes posterior in deep neural networks really?},
  author={Wenzel, Florian and Roth, Kevin and Veeling, Bastiaan S and {\'S}wi{\k{a}}tkowski, Jakub and Tran, Linh and Mandt, Stephan and Snoek, Jasper and Salimans, Tim and Jenatton, Rodolphe and Nowozin, Sebastian},
  journal={arXiv preprint arXiv:2002.02405},
  year={2020}
}

@article{lazaro2010marginalized,
  title={Marginalized neural network mixtures for large-scale regression},
  author={L{\'a}zaro-Gredilla, Miguel and Figueiras-Vidal, An{\'\i}bal R},
  journal={IEEE transactions on neural networks},
  volume={21},
  number={8},
  pages={1345--1351},
  year={2010},
  publisher={IEEE}
}

@article{abe2022deep,
  title={Deep ensembles work, but are they necessary?},
  author={Abe, Taiga and Buchanan, Estefany Kelly and Pleiss, Geoff and Zemel, Richard and Cunningham, John P},
  journal={Advances in Neural Information Processing Systems},
  volume={35},
  pages={33646--33660},
  year={2022}
}

@article{fort2019deep,
  title={Deep ensembles: A loss landscape perspective},
  author={Fort, Stanislav and Hu, Huiyi and Lakshminarayanan, Balaji},
  journal={arXiv preprint arXiv:1912.02757},
  year={2019}
}

@article{harrison2024variational,
  title={Variational Bayesian last layers},
  author={Harrison, James and Willes, John and Snoek, Jasper},
  journal={arXiv preprint arXiv:2404.11599},
  year={2024}
}

@misc{ucirepo,
  author       = {Kelly, Markelle and Longjohn, Rachel and Nottingham, Kolby},
  title        = {{The UCI Machine Learning Repository}},
  howpublished = {\url{https://archive.ics.uci.edu}},
  note         = {University of California, Irvine}
}

@article{an2020deep,
  title={Deep ensemble learning for Alzheimer's disease classification},
  author={An, Ning and Ding, Huitong and Yang, Jiaoyun and Au, Rhoda and Ang, Ting FA},
  journal={Journal of biomedical informatics},
  volume={105},
  pages={103411},
  year={2020},
  publisher={Elsevier}
}

@inproceedings{lee2021sunrise,
  title={Sunrise: A simple unified framework for ensemble learning in deep reinforcement learning},
  author={Lee, Kimin and Laskin, Michael and Srinivas, Aravind and Abbeel, Pieter},
  booktitle={International conference on machine learning},
  pages={6131--6141},
  year={2021},
  organization={PMLR}
}

@article{wu2020deep,
  title={Deep ensemble reinforcement learning with multiple deep deterministic policy gradient algorithm},
  author={Wu, Junta and Li, Huiyun},
  journal={Mathematical Problems in Engineering},
  volume={2020},
  number={1},
  pages={4275623},
  year={2020},
  publisher={Wiley Online Library}
}

@article{seligmann2023beyond,
  title={Beyond deep ensembles: A large-scale evaluation of bayesian deep learning under distribution shift},
  author={Seligmann, Florian and Becker, Philipp and Volpp, Michael and Neumann, Gerhard},
  journal={Advances in Neural Information Processing Systems},
  volume={36},
  pages={29372--29405},
  year={2023}
}

@article{osband2018randomized,
  title={Randomized prior functions for deep reinforcement learning},
  author={Osband, Ian and Aslanides, John and Cassirer, Albin},
  journal={Advances in neural information processing systems},
  volume={31},
  year={2018}
}

@inproceedings{pearce2020uncertainty,
  title={Uncertainty in neural networks: Approximately bayesian ensembling},
  author={Pearce, Tim and Leibfried, Felix and Brintrup, Alexandra},
  booktitle={International conference on artificial intelligence and statistics},
  pages={234--244},
  year={2020},
  organization={PMLR}
}

@article{d2021repulsive,
  title={Repulsive deep ensembles are bayesian},
  author={D'Angelo, Francesco and Fortuin, Vincent},
  journal={Advances in Neural Information Processing Systems},
  volume={34},
  pages={3451--3465},
  year={2021}
}

@article{quinonero2005unifying,
  title={A unifying view of sparse approximate Gaussian process regression},
  author={Quinonero-Candela, Joaquin and Rasmussen, Carl Edward},
  journal={Journal of machine learning research},
  volume={6},
  number={Dec},
  pages={1939--1959},
  year={2005}
}

@inproceedings{wilson2013gaussian,
  title={Gaussian process kernels for pattern discovery and extrapolation},
  author={Wilson, Andrew and Adams, Ryan},
  booktitle={International conference on machine learning},
  pages={1067--1075},
  year={2013},
  organization={PMLR}
}

@article{cho2009kernel,
  title={Kernel methods for deep learning},
  author={Cho, Youngmin and Saul, Lawrence},
  journal={Advances in neural information processing systems},
  volume={22},
  year={2009}
}

@inproceedings{calandra2016manifold,
  title={Manifold Gaussian processes for regression},
  author={Calandra, Roberto and Peters, Jan and Rasmussen, Carl Edward and Deisenroth, Marc Peter},
  booktitle={2016 International joint conference on neural networks (IJCNN)},
  pages={3338--3345},
  year={2016},
  organization={IEEE}
}

@inproceedings{bui2016deep,
  title={Deep Gaussian processes for regression using approximate expectation propagation},
  author={Bui, Thang and Hern{\'a}ndez-Lobato, Daniel and Hernandez-Lobato, Jose and Li, Yingzhen and Turner, Richard},
  booktitle={International conference on machine learning},
  pages={1472--1481},
  year={2016},
  organization={PMLR}
}

@article{snelson2005sparse,
  title={Sparse Gaussian processes using pseudo-inputs},
  author={Snelson, Edward and Ghahramani, Zoubin},
  journal={Advances in neural information processing systems},
  volume={18},
  year={2005}
}

@inproceedings{titsias2009variational,
  title={Variational learning of inducing variables in sparse Gaussian processes},
  author={Titsias, Michalis},
  booktitle={Artificial intelligence and statistics},
  pages={567--574},
  year={2009},
  organization={PMLR}
}

@article{lee2017deep,
  title={Deep neural networks as gaussian processes},
  author={Lee, Jaehoon and Bahri, Yasaman and Novak, Roman and Schoenholz, Samuel S and Pennington, Jeffrey and Sohl-Dickstein, Jascha},
  journal={arXiv preprint arXiv:1711.00165},
  year={2017}
}

@inproceedings{rudner2018connection,
  title={On the connection between neural processes and gaussian processes with deep kernels},
  author={Rudner, Tim GJ and Fortuin, Vincent and Teh, Yee Whye and Gal, Yarin},
  booktitle={Workshop on Bayesian Deep Learning, NeurIPS},
  pages={14},
  year={2018}
}

@article{khan2019approximate,
  title={Approximate inference turns deep networks into Gaussian processes},
  author={Khan, Mohammad Emtiyaz and Immer, Alexander and Abedi, Ehsan and Korzepa, Maciej},
  journal={Advances in neural information processing systems},
  volume={32},
  year={2019}
}

@article{andrieu2003introduction,
  title={An introduction to MCMC for machine learning},
  author={Andrieu, Christophe and De Freitas, Nando and Doucet, Arnaud and Jordan, Michael I},
  journal={Machine learning},
  volume={50},
  number={1},
  pages={5--43},
  year={2003},
  publisher={Springer}
}

@article{hoffman2014no,
  title={The No-U-Turn sampler: adaptively setting path lengths in Hamiltonian Monte Carlo.},
  author={Hoffman, Matthew D and Gelman, Andrew and others},
  journal={J. Mach. Learn. Res.},
  volume={15},
  number={1},
  pages={1593--1623},
  year={2014}
}

@inproceedings{chen2014stochastic,
  title={Stochastic gradient hamiltonian monte carlo},
  author={Chen, Tianqi and Fox, Emily and Guestrin, Carlos},
  booktitle={International conference on machine learning},
  pages={1683--1691},
  year={2014},
  organization={PMLR}
}

@inproceedings{welling2011bayesian,
  title={Bayesian learning via stochastic gradient Langevin dynamics},
  author={Welling, Max and Teh, Yee W},
  booktitle={Proceedings of the 28th international conference on machine learning (ICML-11)},
  pages={681--688},
  year={2011}
}

@article{shridhar2019comprehensive,
  title={A comprehensive guide to bayesian convolutional neural network with variational inference},
  author={Shridhar, Kumar and Laumann, Felix and Liwicki, Marcus},
  journal={arXiv preprint arXiv:1901.02731},
  year={2019}
}

@inproceedings{coker2022wide,
  title={Wide mean-field bayesian neural networks ignore the data},
  author={Coker, Beau and Bruinsma, Wessel P and Burt, David R and Pan, Weiwei and Doshi-Velez, Finale},
  booktitle={International Conference on Artificial Intelligence and Statistics},
  pages={5276--5333},
  year={2022},
  organization={PMLR}
}

@article{sun2019functional,
  title={Functional variational Bayesian neural networks},
  author={Sun, Shengyang and Zhang, Guodong and Shi, Jiaxin and Grosse, Roger},
  journal={arXiv preprint arXiv:1903.05779},
  year={2019}
}

@InProceedings{pmlr-v286-cinquin25a,
  title = 	 {Well-Defined Function-Space Variational Inference in Bayesian Neural Networks via Regularized KL-Divergence},
  author =       {Cinquin, Tristan and Bamler, Robert},
  booktitle = 	 {Proceedings of the Forty-first Conference on Uncertainty in Artificial Intelligence},
  pages = 	 {752--776},
  year = 	 {2025},
  editor = 	 {Chiappa, Silvia and Magliacane, Sara},
  volume = 	 {286},
  series = 	 {Proceedings of Machine Learning Research},
  month = 	 {21--25 Jul},
  publisher =    {PMLR},
}

@inproceedings{shah2014student,
  title={Student-t processes as alternatives to Gaussian processes},
  author={Shah, Amar and Wilson, Andrew and Ghahramani, Zoubin},
  booktitle={Artificial intelligence and statistics},
  pages={877--885},
  year={2014},
  organization={PMLR}
}

@article{fiedler2023improved,
  title={Improved uncertainty quantification for neural networks with bayesian last layer},
  author={Fiedler, Felix and Lucia, Sergio},
  journal={IEEE Access},
  volume={11},
  pages={123149--123160},
  year={2023},
  publisher={IEEE}
}

@inproceedings{stoyanova2023toward,
  title={Toward robust uncertainty estimation with random activation functions},
  author={Stoyanova, Yana and Ghandi, Soroush and Tavakol, Maryam},
  booktitle={Proceedings of the AAAI Conference on Artificial Intelligence},
  volume={37},
  number={12},
  pages={15152--15160},
  year={2023}}

@inproceedings{jain2020maximizing,
  title={Maximizing overall diversity for improved uncertainty estimates in deep ensembles},
  author={Jain, Siddhartha and Liu, Ge and Mueller, Jonas and Gifford, David},
  booktitle={Proceedings of the AAAI conference on artificial intelligence},
  volume={34},
  number={04},
  pages={4264--4271},
  year={2020}
}

@article{reeve2018diversity,
  title={Diversity and degrees of freedom in regression ensembles},
  author={Reeve, Henry WJ and Brown, Gavin},
  journal={Neurocomputing},
  volume={298},
  pages={55--68},
  year={2018},
  publisher={Elsevier}
}

@article{rame2021dice,
  title={Dice: Diversity in deep ensembles via conditional redundancy adversarial estimation},
  author={Rame, Alexandre and Cord, Matthieu},
  journal={arXiv preprint arXiv:2101.05544},
  year={2021}
}

@article{nam2021diversity,
  title={Diversity matters when learning from ensembles},
  author={Nam, Giung and Yoon, Jongmin and Lee, Yoonho and Lee, Juho},
  journal={Advances in neural information processing systems},
  volume={34},
  pages={8367--8377},
  year={2021}
}

@article{zaidi2021neural,
  title={Neural ensemble search for uncertainty estimation and dataset shift},
  author={Zaidi, Sheheryar and Zela, Arber and Elsken, Thomas and Holmes, Chris C and Hutter, Frank and Teh, Yee},
  journal={Advances in Neural Information Processing Systems},
  volume={34},
  pages={7898--7911},
  year={2021}
}

@book{sutton1998reinforcement,
  title={Reinforcement learning: An introduction},
  author={Sutton, Richard S and Barto, Andrew G and others},
  volume={1},
  number={1},
  year={1998},
  publisher={MIT press Cambridge}
}

@article{levine2020offline,
  title={Offline reinforcement learning: Tutorial, review, and perspectives on open problems},
  author={Levine, Sergey and Kumar, Aviral and Tucker, George and Fu, Justin},
  journal={arXiv preprint arXiv:2005.01643},
  year={2020}
}

@inproceedings{deisenroth2011pilco,
  title={PILCO: A model-based and data-efficient approach to policy search},
  author={Deisenroth, Marc and Rasmussen, Carl E},
  booktitle={Proceedings of the 28th International Conference on machine learning (ICML-11)},
  pages={465--472},
  year={2011}
}

@article{arulkumaran2017deep,
  title={Deep reinforcement learning: A brief survey},
  author={Arulkumaran, Kai and Deisenroth, Marc Peter and Brundage, Miles and Bharath, Anil Anthony},
  journal={IEEE signal processing magazine},
  volume={34},
  number={6},
  pages={26--38},
  year={2017},
  publisher={IEEE}
}

@article{raissi2019physics,
  title={Physics-informed neural networks: A deep learning framework for solving forward and inverse problems involving nonlinear partial differential equations},
  author={Raissi, Maziar and Perdikaris, Paris and Karniadakis, George E},
  journal={Journal of Computational physics},
  volume={378},
  pages={686--707},
  year={2019},
  publisher={Elsevier}
}

@book{gelman1995bayesian,
  title={Bayesian data analysis},
  author={Gelman, Andrew and Carlin, John B and Stern, Hal S and Rubin, Donald B},
  year={1995},
  publisher={Chapman and Hall/CRC}
}

@article{jeffreys1946invariant,
  title={An invariant form for the prior probability in estimation problems},
  author={Jeffreys, Harold},
  journal={Proceedings of the Royal Society of London. Series A. Mathematical and Physical Sciences},
  volume={186},
  number={1007},
  pages={453--461},
  year={1946},
  publisher={The Royal Society London}
}
\bibliographystyle{plainnat}

%%%%%%%%%%%%%%%%%%%%%%%%%%%%%%%%%%%%%%%%%%%%%%%%%%%%%%%%%%%%%%%%%%%%%%%%%%%%%%%
%%%%%%%%%%%%%%%%%%%%%%%%%%%%%%%%%%%%%%%%%%%%%%%%%%%%%%%%%%%%%%%%%%%%%%%%%%%%%%%
% APPENDIX
%%%%%%%%%%%%%%%%%%%%%%%%%%%%%%%%%%%%%%%%%%%%%%%%%%%%%%%%%%%%%%%%%%%%%%%%%%%%%%%
%%%%%%%%%%%%%%%%%%%%%%%%%%%%%%%%%%%%%%%%%%%%%%%%%%%%%%%%%%%%%%%%%%%%%%%%%%%%%%%

\newpage
\appendix
\onecolumn
\section{Appendix}
\label{sec:appendix}

% \subsection{Model derivations}

\subsection{Jeffreys prior for the noise variance}
\label{subsub:jeffreys}

We derive the Jeffreys prior for the Gaussian noise variance $\sigma^2_{\varepsilon}$ in the
Bayesian linear model used in Step~2.
Consider the observation model
\[
\mathbf{y}\mid \boldsymbol{\beta},\sigma^2_{\varepsilon}
\sim
\mathcal{N}\!\big(\boldsymbol{\Phi}\boldsymbol{\beta},\;\sigma^2_{\varepsilon}\mathbf{I}_N\big),
\]
where $\boldsymbol{\Phi}\in\mathbb{R}^{N\times H}$ is the ensemble-induced design
matrix.
The log-likelihood of $\sigma^2_{\varepsilon}$ is
\[
\ell(\sigma^2_{\varepsilon})
=
-\frac{N}{2}\log\sigma^2_{\varepsilon}
-\frac{1}{2\sigma^2_{\varepsilon}}
\|\mathbf{y}-\boldsymbol{\Phi}\boldsymbol{\beta}\|^2
+ C,
\]
where $C$ denotes a constant independent of $\sigma^2_{\varepsilon}$.
Differentiating twice with respect to $\sigma^2_{\varepsilon}$ yields
\[
\frac{\partial^2 \ell}{\partial (\sigma^2_{\varepsilon})^2}
=
\frac{N}{2}\,\frac{1}{(\sigma^2_{\varepsilon})^2}
-
\frac{\|\mathbf{y}-\boldsymbol{\Phi}\boldsymbol{\beta}\|^2}{(\sigma^2_{\varepsilon})^3}.
\]
Taking the negative expectation under
$\mathbf{y}\sim\mathcal{N}(\boldsymbol{\Phi}\boldsymbol{\beta},\sigma^2_{\varepsilon}\mathbf{I}_N)$
gives the Fisher information
\[
\mathcal{I}(\sigma^2_{\varepsilon})
=
\frac{N}{2}\,\frac{1}{(\sigma^2_{\varepsilon})^2}.
\]
By definition, the Jeffreys prior is proportional to the square root of the
Fisher information, yielding
\[
p(\sigma^2_{\varepsilon})\;\propto\;\sqrt{\mathcal{I}(\sigma^2_{\varepsilon})}\;=\;\frac{1}{\sigma^2_{\varepsilon}}.
\]

% Kernel derivation for the induced GP (conditional on \sigma^2)

\subsection{GP connection}
\label{subsub:gp_connection}

Recall 
\begin{equation*}
f(x) = \boldsymbol{\phi}(x)^\top \boldsymbol{\beta},
\qquad
\boldsymbol{\beta}\mid\sigma^2_{\varepsilon} \sim
\mathcal{N}\!\big(\boldsymbol{\mu}_0,\;\sigma^2_{\varepsilon}\boldsymbol{\Lambda}_0^{-1}\big).
\end{equation*}

The conditional mean function is
\begin{align*}
m(x)
&= \mathbb{E}\!\big[f(x)\mid\sigma^2_{\varepsilon}\big] \\
&= \mathbb{E}\!\big[\boldsymbol{\phi}(x)^\top\boldsymbol{\beta}\mid\sigma^2_{\varepsilon}\big] \\
&= \boldsymbol{\phi}(x)^\top \boldsymbol{\mu}_0.
\end{align*}

The conditional covariance (kernel) is given by
\begin{align*}
k(x,x'\mid\sigma^2_{\varepsilon})
&= \mathrm{Cov}\!\big(f(x), f(x') \mid \sigma^2_{\varepsilon}\big) \\
&= \mathbb{E}\!\Big[
\big(f(x)-m(x)\big)\big(f(x')-m(x')\big)
\,\Big|\, \sigma^2_{\varepsilon}
\Big] \\
&= \mathbb{E}\!\Big[
\big(\boldsymbol{\phi}(x)^\top\boldsymbol{\beta}
      - \boldsymbol{\phi}(x)^\top\boldsymbol{\mu}_0\big)
\big(\boldsymbol{\phi}(x')^\top\boldsymbol{\beta}
      - \boldsymbol{\phi}(x')^\top\boldsymbol{\mu}_0\big)
\,\Big|\, \sigma^2_{\varepsilon}
\Big] \\
&= \mathbb{E}\!\Big[
\boldsymbol{\phi}(x)^\top(\boldsymbol{\beta}-\boldsymbol{\mu}_0)
(\boldsymbol{\beta}-\boldsymbol{\mu}_0)^\top
\boldsymbol{\phi}(x')
\,\Big|\, \sigma^2_{\varepsilon}
\Big] \\
&= \boldsymbol{\phi}(x)^\top
\mathbb{E}\!\Big[
(\boldsymbol{\beta}-\boldsymbol{\mu}_0)
(\boldsymbol{\beta}-\boldsymbol{\mu}_0)^\top
\,\Big|\, \sigma^2_{\varepsilon}
\Big]
\boldsymbol{\phi}(x') \\
&= \boldsymbol{\phi}(x)^\top
\mathrm{Cov}(\boldsymbol{\beta}\mid\sigma^2_{\varepsilon})
\boldsymbol{\phi}(x') \\
&= \boldsymbol{\phi}(x)^\top
\big(\sigma^2_{\varepsilon} \boldsymbol{\Lambda}_0^{-1}\big)
\boldsymbol{\phi}(x') \\
&= \sigma^2_{\varepsilon}\,\boldsymbol{\phi}(x)^\top
\boldsymbol{\Lambda}_0^{-1}
\boldsymbol{\phi}(x').
\end{align*}

Writing
\(
\boldsymbol{\phi}(x) = [f_1(x),\dots,f_H(x)]^\top
\)
and
\(
\boldsymbol{\Lambda}_0^{-1}
= \big[(\boldsymbol{\Lambda}_0^{-1})_{hh'}\big]_{h,h'=1}^H,
\)
the kernel can be expressed as
\begin{equation*}
k(x,x'\mid\sigma^2_{\varepsilon})
= \sigma^2_{\varepsilon}
\sum_{h=1}^H
\sum_{h'=1}^H
(\boldsymbol{\Lambda}_0^{-1})_{hh'}
\, f_h(x)\, f_{h'}(x').
\end{equation*}

\paragraph{Inner-product (finite-rank) form.}
Let $\boldsymbol{\Lambda}_0^{-1/2}$ denote any symmetric square root of
$\boldsymbol{\Lambda}_0^{-1}$ and define the transformed feature map
$\tilde{\boldsymbol{\phi}}(x)\triangleq \boldsymbol{\Lambda}_0^{-1/2}\boldsymbol{\phi}(x)$.
With this notation, the kernel has the inner-product representation
\[
k(x,x'\mid\sigma^2_{\varepsilon})
= \sigma^2_{\varepsilon}\,\tilde{\boldsymbol{\phi}}(x)^\top \tilde{\boldsymbol{\phi}}(x').
\]
Consequently, for any collection of inputs $\{x_i\}_{i=1}^N$, the associated
kernel matrix can be written as $K=\sigma^2_{\varepsilon}\tilde{\Phi}\tilde{\Phi}^\top$ with
$\tilde{\Phi}\in\mathbb{R}^{N\times H}$, implying that $\mathrm{rank}(K)\le H$
independently of $N$. The induced Gaussian Process is therefore finite-rank, with
sample paths that are almost surely confined to the $H$-dimensional linear subspace
$\mathrm{span}\{f_1,\dots,f_H\}$. In contrast to nondegenerate (full-rank) kernels such as the squared-exponential or Matern kernels \cite{williams2006gaussian}, the function-space complexity of the proposed model is explicitly controlled by the ensemble size $H$.

\subsection{Isotropic prior over ensemble coefficients}
\label{subsub:beta_isotropic}

In our experiments, we adopt an isotropic Gaussian prior over the ensemble
coefficients by setting
\begin{equation*}
\boldsymbol{\Lambda}_0 = \sigma_{\beta}^{-2}\mathbf{I}_H,
\end{equation*}
which corresponds to
\(
\boldsymbol{\beta}\mid\sigma^2_{\varepsilon} \sim
\mathcal{N}(\boldsymbol{\mu}_0,\;\sigma^2_{\varepsilon}\sigma_{\beta}^2 \mathbf{I}_H).
\)
Under this choice, all ensemble-induced basis functions are a priori treated
symmetrically and assigned the same variance, reflecting the absence of
additional structural assumptions about the relative importance of individual
ensemble members.

Substituting $\boldsymbol{\Lambda}_0^{-1} = \sigma_{\beta}^2 \mathbf{I}_H$ into the kernel expression derived in the previous section yields
\begin{align*}
k(x,x'\mid\sigma^2_{\varepsilon})
&= \sigma^2_{\varepsilon}\,\boldsymbol{\phi}(x)^\top
\boldsymbol{\Lambda}_0^{-1}
\boldsymbol{\phi}(x') \\
&= \sigma^2_{\varepsilon} \sigma_{\beta}^2
\boldsymbol{\phi}(x)^\top \boldsymbol{\phi}(x') \\
&= \sigma^2_{\varepsilon} \sigma_{\beta}^2
\sum_{h=1}^H f_h(x)\,f_h(x').
\end{align*}
This isotropic prior offers several practical and conceptual advantages.
First, it avoids introducing additional biases that would favor
specific ensemble members, allowing uncertainty to be driven primarily by the
diversity of the learned functions themselves. Second, the single scale
parameter $\sigma_{\beta}^2$ controls the overall strength of the prior in the
ensemble-induced function space, enabling efficient empirical Bayes tuning via
the marginal likelihood without increasing model complexity. Finally, because the dimensionality of the induced feature space is small (equal to the ensemble
size $H$), inference remains computationally inexpensive while retaining a
principled Bayesian interpretation equivalent to finite-rank Gaussian Process
regression.

%%%%%%%%%%%%%%%%%%%%%%%%%%%%%%%%%%%%%%%%%%%%%%%%%%%%%%%%%%%%%%%%%%%%%%%%%%%%%%%

\subsection{BDKN inference}
\label{subsec:bdkn_algo}

This section discuss the posterior updates, marginal likelihood optimization, and inference algorithm used in the Bayesian inference procedure of BDKN.

\subsubsection*{Posterior updates}

Given the design matrix $\boldsymbol{\Phi}$ and targets
$\mathbf{y}$, conjugacy is maintained and the posterior factorizes as \(
p(\boldsymbol{\beta},\sigma^2_{\varepsilon}\mid \mathbf{y},\boldsymbol{\Phi})
= p(\boldsymbol{\beta}\mid\sigma^2_{\varepsilon},\mathbf{y},\boldsymbol{\Phi})\;
  p(\sigma^2_{\varepsilon}\mid \mathbf{y},\boldsymbol{\Phi}).
\)
The conditional posterior over coefficients is Gaussian,
\(
\boldsymbol{\beta}\mid\sigma^2_{\varepsilon},\mathbf{y},\boldsymbol{\Phi}
\sim \mathcal{N}\!\big(\boldsymbol{\mu}_n,\; \sigma^2_{\varepsilon}\boldsymbol{\Lambda}_n^{-1}\big),
\)
with
\begin{align}
\label{eq:mu_n}
\boldsymbol{\mu}_n &= \boldsymbol{\Lambda}_n^{-1}
\big(\boldsymbol{\Lambda}_0\boldsymbol{\mu}_0 + \boldsymbol{\Phi}^\top \mathbf{y}\big),\\
\boldsymbol{\Lambda}_n &= \boldsymbol{\Lambda}_0 + \boldsymbol{\Phi}^\top\boldsymbol{\Phi}.
\end{align}
Under the Jeffreys prior on $\sigma^2_{\varepsilon}$, the marginal posterior over $\sigma^2_{\varepsilon}$ is \(
\sigma^2_{\varepsilon}\mid \mathbf{y},\boldsymbol{\Phi}\sim \mathrm{Inv\text{-}Gamma}(a_n,b_n),
\)
where
\begin{align}
\label{eq:b_n}
a_n &= \frac{N}{2},\\
b_n &= \frac{1}{2}\!\left(
\mathbf{y}^\top\mathbf{y}
+ \boldsymbol{\mu}_0^\top\boldsymbol{\Lambda}_0\boldsymbol{\mu}_0
- \boldsymbol{\mu}_n^\top\boldsymbol{\Lambda}_n\boldsymbol{\mu}_n
\right).
\end{align}

\subsubsection*{Marginal likelihood}

The hyperparameter $\bm{\Lambda}_0$, specifically $\sigma^2_{\beta}$, is optimized by an empirical Bayes estimate, maximizing the marginal likelihood in a closed form.
% When adopting Jeffreys' prior for the noise variance, the marginal likelihood is
% defined only up to an additive constant. 
% due to the improper normalization of the prior.
Specifically, one maximizes the log marginal likelihood, written as
\begin{equation}
\log p(\mathbf{y}\mid \boldsymbol{\Phi})
=
-a_n\log b_n
+\frac{1}{2}\log\lvert\boldsymbol{\Lambda}_0\rvert
-\frac{1}{2}\log\lvert\boldsymbol{\Lambda}_n\rvert
+ C,
\label{eq:log_marginal_likelihood}
\end{equation}
% \todo{Is this correct? Why does this have beta?}
% \commentt{this beta is inverse gamma parameter, I will change to a and b instead of alpha and beta to avoid confusion}
where $C$ denotes a constant independent of the hyperparameters. 
% The above marginal likelihood is used to optimize the hyperparameters of the prior on $\boldsymbol{\beta}$ by evidence maximization, resulting in an empirical Bayes estimate. 
Algorithm~\ref{alg:bdkn} outlines the main steps of BDKN.

\begin{algorithm}[H]
\caption{Bayesian Deep Kernel Networks (BDKN)}
\label{alg:bdkn}
\begin{algorithmic}[1]
\STATE \textbf{Inputs:} training set $\mathcal{D}_{\mathrm{tr}}=\{(\bx_i,y_i)\}_{i=1}^{N}$, ensemble size $H$, initial prior variance $\sigma_{\beta}^2 = 1$
\STATE \textbf{Outputs:} feature maps $\{\phi_h\}_{h=1}^H$ and posterior predictive distribution $p(y_* \mid \bx_*,\mathcal{D}_{\mathrm{tr}},\{\phi_h\})$

\vspace{0.25em}
\STATE \textbf{Step 1: Learn feature maps via independent training.}
\STATE Train $H$ neural networks independently on $\mathcal{D}_{\mathrm{tr}}$ to obtain $\{\phi_h(\cdot;\btheta_h)\}_{h=1}^H$.

\vspace{0.25em}
\STATE \textbf{Step 2: Bayesian inference over learned features.}
\STATE Construct the design matrix $\boldsymbol{\Phi}$ using $\{\phi_h\}_{h=1}^H$.
\STATE Optimize $\sigma_{\beta}^2$ by maximizing the log marginal likelihood in \eqref{eq:log_marginal_likelihood}.
\STATE Compute posterior parameters $(\boldsymbol{\mu}_n,\boldsymbol{\Lambda}_n,a_n,b_n)$ using  \eqref{eq:mu_n}-\eqref{eq:b_n}
\STATE Obtain the posterior predictive distribution using \eqref{eq:posterior prediction}.

\end{algorithmic}
\end{algorithm}

\subsection{Experiments details}
\label{subsec:experiments_details}

\subsubsection{Data}
\label{subsub:data}

We evaluated all methods on standard UCI regression benchmarks. Table~\ref{tab:dataset_stats} summarizes the key characteristics of each dataset, including the total number of samples, input dimensionality, number of random splits, and the corresponding train-test sizes. Except for Song, experiments use a 90\%/10\% train-test split repeated across multiple random splits as reported in the table; Song uses the fixed train-test split shown in the table. In all experiments, input features are standardized using statistics computed on the training data, resulting in zero-mean and unit-variance inputs.

\begin{table}[H]
\centering
\setlength{\tabcolsep}{4pt}
\renewcommand{\arraystretch}{1.1}
\caption{Dataset statistics for standard regression benchmarks.}
\label{tab:dataset_stats}

\begin{tabular}{lccccc}
\toprule
Dataset
& Total Samples
& \# Features
& \# Splits
& Train size
& Test size \\
\midrule
Boston
& 506
& 13
& 20
& 455
& 51 \\

Concrete
& 1030
& 8
& 20
& 927
& 103 \\

Energy
& 768
& 8
& 20
& 691
& 77 \\

Kin8nm
& 8192
& 8
& 20
& 7373
& 819 \\

Naval
& 11934
& 16
& 20
& 10741
& 1193 \\

Power
& 9568
& 4
& 20
& 8611
& 957 \\

Protein
& 45730
& 9
& 5
& 41157
& 4573 \\

Sarcos
& 39449
& 21
& 20
& 35000
& 4449 \\

Song
& 515345
& 90
& 1
& 463715
& 51630 \\

Wine
& 1599
& 11
& 20
& 1439
& 160 \\

Yacht
& 308
& 6
& 20
& 277
& 31 \\
\bottomrule
\end{tabular}
\end{table}

\subsubsection{Settings}
\label{subsub:settings}

Table \ref{tab:setting2_hparams} summarizes the optimization hyperparameters used in the first experimental setting, which corresponds to a high learning-rate configuration. Following the lightweight training regime commonly adopted in DE benchmarks, all models are trained with a learning rate of 0.1 for the full budget of 40 epochs on all datasets. Activation functions follow standard choices for each model class (ReLU for BDKN, DE, and VBLL; Leaky ReLU for BLL, LD-BLL, and MFVI-BNN), GP uses an RBF kernel, and DKL uses a ReLU feature extractor with an RBF GP layer. Results corresponding to this configuration are referred to as Setting~1 throughout the paper.

\begin{table}[H]
\centering
\small
\setlength{\tabcolsep}{3pt}
\renewcommand{\arraystretch}{1.05}

\caption{Learning rates and maximum number of epochs for regression tasks (Setting~1 high learning rate). All neural-network models use hidden dimensions \(50,50\).}
\label{tab:setting2_hparams}

\begin{tabular}{lcccccccc}
\toprule
Method & BDKN & DE & BLL & LDBLL & MFVI-BNN & GP & DKL & VBLL \\
Activation/Kernel & ReLU & ReLU & LReLU & LReLU & LReLU & RBF & ReLU/RBF & ReLU \\
\midrule
Boston   & 1e-1, 40 & 1e-1, 40 & 1e-1, 40 & 1e-1, 40 & 1e-1, 40 & 1e-1, 40 & 1e-1, 40 & 1e-1, 40 \\
Concrete & 1e-1, 40 & 1e-1, 40 & 1e-1, 40 & 1e-1, 40 & 1e-1, 40 & 1e-1, 40 & 1e-1, 40 & 1e-1, 40 \\
Energy   & 1e-1, 40 & 1e-1, 40 & 1e-1, 40 & 1e-1, 40 & 1e-1, 40 & 1e-1, 40 & 1e-1, 40 & 1e-1, 40 \\
Kin8nm   & 1e-1, 40 & 1e-1, 40 & 1e-1, 40 & 1e-1, 40 & 1e-1, 40 & 1e-1, 40 & 1e-1, 40 & 1e-1, 40 \\
Naval    & 1e-1, 40 & 1e-1, 40 & 1e-1, 40 & 1e-1, 40 & 1e-1, 40 & 1e-1, 40 & 1e-1, 40 & 1e-1, 40 \\
Power    & 1e-1, 40 & 1e-1, 40 & 1e-1, 40 & 1e-1, 40 & 1e-1, 40 & 1e-1, 40 & 1e-1, 40 & 1e-1, 40 \\
Protein  & 1e-1, 40 & 1e-1, 40 & 1e-1, 40 & 1e-1, 40 & 1e-1, 40 & N/A & 1e-1, 40 & 1e-1, 40 \\
Sarcos   & 1e-1, 40 & 1e-1, 40 & 1e-1, 40 & 1e-1, 40 & 1e-1, 40 & N/A & 1e-1, 40 & 1e-1, 40 \\
Song     & 1e-1, 40 & 1e-1, 40 & 1e-1, 40 & 1e-1, 40 & 1e-1, 40 & N/A & 1e-1, 40 & 1e-1, 40 \\
Wine     & 1e-1, 40 & 1e-1, 40 & 1e-1, 40 & 1e-1, 40 & 1e-1, 40 & 1e-1, 40 & 1e-1, 40 & 1e-1, 40 \\
Yacht    & 1e-1, 40 & 1e-1, 40 & 1e-1, 40 & 1e-1, 40 & 1e-1, 40 & 1e-1, 40 & 1e-1, 40 & 1e-1, 40 \\
\bottomrule
\end{tabular}
\end{table}

Table \ref{tab:setting1_hparams} summarizes the optimization hyperparameters used in the second experimental setting, which corresponds to a lower learning-rate configuration. This setting follows the benchmark protocols proposed by Watson et al.~\cite{watson2021latent}, using dataset- and method-specific learning rates and training budgets as reported in the table. For BDKN, DE, DKL, and VBLL, models are trained for the full budget of 500 epochs on all datasets. For BLL, LD-BLL, and MFVI-BNN, the effective number of training epochs is selected via validation: for each split, 80\% of the training data are used for optimization and the remaining 20\% are used as a validation set to determine the epoch with the best validation performance. The model is then retrained on the full training set using this selected number of epochs. For GP regression, validation-based early stopping is computationally expensive; therefore, we employ a marginal-likelihood--based stopping criterion and terminate optimization when the relative decrease in the average marginal likelihood over the most recent \(\kappa\) iterations, compared to the average over the preceding \(\kappa\) iterations, falls below a threshold \(\rho\), following a procedure similar to that used in prior work \cite{watson2021latent}. For all datasets, we set \(\kappa = 11\) and \(\rho = 10^{-4}\). Apart from the learning rates and maximum epochs specified in Table \ref{tab:setting1_hparams}, all other experimental details are unchanged. Results obtained under this configuration are referred to as Setting~2 throughout the paper.

\begin{table}[H]
\centering
\footnotesize
\setlength{\tabcolsep}{3pt}
\renewcommand{\arraystretch}{1.05}

\caption{Learning rates and maximum number of epochs for regression tasks (Setting~2 low learning rate). All neural-network models use hidden dimensions \(50,50\), except on Sarcos where BLL, LDBLL, and MFVI-BNN use hidden dimensions \(50,200,200\).}
\label{tab:setting1_hparams}

\begin{tabular}{lcccccccc}
\toprule
Method & BDKN & DE & BLL & LDBLL & MFVI-BNN & GP & DKL & VBLL \\
Activation/Kernel & ReLU & ReLU & LReLU & LReLU & LReLU & RBF & ReLU/RBF & ReLU \\
\midrule
Boston   & 1e-3, 500 & 1e-3, 500 & 1e-3, 3000 & 1e-3, 4000 & 1e-3, 10000 & 1e-2, 1000 & 1e-3, 500 & 1e-3, 500 \\
Concrete & 1e-3, 500 & 1e-3, 500 & 1e-3, 3000 & 1e-3, 4000 & 1e-3, 15000 & 1e-2, 1000 & 1e-3, 500 & 1e-3, 500 \\
Energy   & 1e-3, 500 & 1e-3, 500 & 1e-3, 8000 & 1e-3, 10000 & 1e-3, 30000 & 1e-2, 2000 & 1e-3, 500 & 1e-3, 500 \\
Kin8nm   & 1e-3, 500 & 1e-3, 500 & 1e-3, 3000 & 1e-3, 5000 & 1e-3, 20000 & 1e-2, 1000 & 1e-3, 500 & 1e-3, 500 \\
Naval    & 1e-3, 500 & 1e-3, 500 & 1e-3, 8000 & 1e-3, 10000 & 1e-3, 100000 & 1e-3, 3000 & 1e-3, 500 & 1e-3, 500 \\
Power    & 1e-3, 500 & 1e-3, 500 & 1e-3, 5000 & 1e-3, 5000 & 1e-3, 20000 & 1e-2, 1000 & 1e-3, 500 & 1e-3, 500 \\
Protein  & 1e-3, 500 & 1e-3, 500 & 1e-3, 2000 & 1e-3, 5000 & 1e-3, 30000 & N/A & 1e-3, 500 & 1e-3, 500 \\
Sarcos   & 1e-3, 500 & 1e-3, 500 & 2e-4, 30000 & 2e-4, 30000 & 5e-3, 10000 & N/A & 1e-3, 500 & 1e-3, 500 \\
Song     & 1e-3, 500 & 1e-3, 500 & 1e-3, 500 & N/A & 1e-3, 500 & N/A & 1e-3, 500 & 1e-3, 500 \\
Wine     & 1e-3, 500 & 1e-3, 500 & 1e-4, 1000 & 1e-4, 1000 & 1e-4, 20000 & 1e-2, 1000 & 1e-3, 500 & 1e-3, 500 \\
Yacht    & 1e-3, 500 & 1e-3, 500 & 1e-3, 8000 & 1e-3, 10000 & 1e-3, 20000 & 1e-2, 2000 & 1e-3, 500 & 1e-3, 500 \\
\bottomrule
\end{tabular}
\end{table}

For both experimental settings, all neural network-based methods use architectures with two fully connected hidden layers of width 50 and are trained with a batch size of 32 using the Adam optimizer, except for BLL, LD-BLL, and MFVI-BNN on Sarcos in the low-learning-rate setting, which use hidden dimensions \(50,200,200\). DKL uses the same ReLU feature extractor architecture as the other neural-network baselines, with a two-dimensional output representation following Wilson et al.~\cite{wilson2016deep}, and places an RBF GP layer on top of the learned representation. For MFVI-BNN, we follow the priors and inference procedure described in \cite{watson2021latent}, using independent \(\mathcal{N}(0, \omega/\sqrt{n_{\mathrm{in}}})\) priors for all network weights, where \(n_{\mathrm{in}}\) denotes the number of input features to the corresponding layer and \(\omega = 4\), and independent \(\mathcal{N}(0,1)\) priors for bias terms. Variational inference is implemented in \texttt{Pyro} using diagonal Gaussian variational distributions, optimized via stochastic variational inference (SVI) with the \texttt{Trace\_ELBO} objective. For validation and evaluation, predictive distributions are estimated using 100 Monte Carlo samples from the variational posterior. VBLL is implemented using the official Python library (\texttt{VBLL}) provided in \cite{harrison2024variational} with the default settings reported in the paper, and GP regression is implemented using \texttt{GPyTorch}.

For both experimental settings, we initialize the prior variance to $\sigma_{\beta}^2 = 1$ and perform a small number of marginal-likelihood optimization steps (5 iterations) using the Adam optimizer with a learning rate of $0.1$, following Algorithm~\ref{alg:bdkn}. We adopt a zero-mean isotropic Gaussian prior over the linear coefficients, i.e., $\boldsymbol{\beta} \sim \mathcal{N}(0, \sigma_{\beta}^2 I)$, which is a standard and commonly used choice in Bayesian last-layer models \cite{watson2021latent}.

\subsubsection{Runtime}
\label{subsub:result}

All experiments were conducted on a workstation running Windows 10 Enterprise (64-bit), equipped with a 12th-generation Intel Core i7-12700 CPU (12 cores, 20 logical processors), 32~GB of RAM, and an NVIDIA RTX A2000 GPU with 12~GB of dedicated memory. The system uses NVMe SSD storage. GPU computations were performed using CUDA~12.8 with NVIDIA driver version 573.44. Tables~\ref{tab:time_Setting1} and~\ref{tab:time_setting2} report the training time (in minutes) for all methods under the two experimental settings. The runtimes correspond to end-to-end training on standard regression benchmarks using the same hardware and implementation framework. Under Setting~1 (high learning rate), BDKN exhibits runtime comparable to standard DE, reflecting that its additional Bayesian aggregation step incurs negligible overhead relative to training the ensemble members. In contrast, fully Bayesian or variational approaches, such as MFVI-BNN and VBLL, incur substantially higher computational costs on larger datasets due to sampling-based inference or variational optimization. GP is efficient on smaller datasets but becomes expensive or infeasible on larger datasets, such as \textit{Protein, Sarcos}, and \textit{Song}.  
Under Setting~2 (low learning rate), overall runtimes increase across all neural network-based methods due to longer training schedules, with MFVI-BNN showing particularly large increases on high-dimensional or large-scale datasets. Despite this, BDKN continues to scale comparably to DE and remains significantly more efficient than fully Bayesian baselines. Entries marked as \textit{N/A} indicate cases where a method could not be evaluated due to GPU memory constraints.

\begin{table}[H]
\centering
\setlength{\tabcolsep}{4pt}
\renewcommand{\arraystretch}{1.1}

\caption{Runtime (minutes) on standard regression benchmarks (Setting 1).}
\label{tab:time_Setting1}

\begin{tabular*}{\textwidth}{@{\extracolsep{\fill}}lcccccccc}
\toprule
Dataset & BDKN & DE & LDBLL & GP & DKL & BLL & MFVI & VBLL \\
\midrule
Boston & 0.76 & 0.66 & 0.36 & 0.10 & 0.67 & 0.06 & 2.06 & 1.43 \\
Concrete & 0.75 & 0.66 & 1.40 & 0.59 & 0.30 & 0.23 & 3.59 & 2.19 \\
Energy & 0.54 & 0.46 & 0.15 & 0.45 & 0.52 & 0.09 & 3.36 & 1.79 \\
Kin8nm & 6.08 & 5.99 & 1.64 & 2.55 & 0.72 & 0.19 & 5.14 & 17.61 \\
Naval & 7.30 & 7.21 & 0.46 & 4.90 & 1.91 & 0.05 & 10.43 & 18.90 \\
Power & 6.40 & 6.32 & 0.24 & 3.91 & 1.29 & 0.02 & 8.17 & 17.68 \\
Protein & 8.47 & 8.45 & 0.44 & N/A & 0.27 & 0.13 & 16.24 & 25.56 \\
Sarcos & 30.04 & 29.95 & 2.52 & N/A & 2.29 & 0.47 & 63.65 & 80.45 \\
Song & 17.51 & 17.50 & N/A & N/A & 15.04 & 0.39 & 37.28 & 99.67 \\
Wine & 2.08 & 1.94 & 0.19 & 1.31 & 1.19 & 0.04 & 5.45 & 6.83 \\
Yacht & 0.61 & 0.45 & 0.32 & 0.16 & 0.89 & 0.15 & 5.86 & 1.33 \\
\bottomrule
\end{tabular*}
\end{table}

\begin{table}[H]
\centering
\setlength{\tabcolsep}{4pt}
\renewcommand{\arraystretch}{1.1}

\caption{Runtime (minutes) on standard regression benchmarks (Setting 2).}
\label{tab:time_setting2}

\begin{tabular*}{\textwidth}{@{\extracolsep{\fill}}lcccccccc}
\toprule
Dataset & BDKN & DE & LDBLL & GP & DKL & BLL & MFVI & VBLL \\
\midrule
Boston & 4.07 & 4.07 & 14.68 & 33.42 & 8.93 & 20.07 & 902.33 & 13.27 \\
Concrete & 8.31 & 8.26 & 11.36 & 30.79 & 14.86 & 10.59 & 829.53 & 28.13 \\
Energy & 6.73 & 6.72 & 113.53 & 14.79 & 11.75 & 29.54 & 3176.77 & 20.19 \\
Kin8nm & 68.56 & 68.54 & 122.19 & 55.82 & 35.90 & 0.94 & 2357.65 & 234.30 \\
Naval & 91.10 & 91.09 & 241.29 & 336.05 & 22.15 & 127.35 & 14625.90 & 302.48 \\
Power & 80.87 & 80.87 & 22.39 & 81.07 & 11.58 & 85.03 & 1960.26 & 421.37 \\
Protein & 111.35 & 111.35 & 106.79 & N/A & 2.70 & 14.54 & 6099.07 & 358.99 \\
Sarcos & 362.98 & 362.98 & 8494.25 & N/A & 19.62 & 669.92 & 7993.33 & 1256.18 \\
Song & 229.29 & 229.28 & N/A & N/A & 78.31 & 5.62 & 488.23 & 765.11 \\
Wine & 12.18 & 12.17 & 4.81 & 19.03 & 7.99 & 34.93 & 2267.03 & 47.35 \\
Yacht & 3.20 & 3.19 & 117.50 & 11.15 & 8.34 & 37.28 & 2031.11 & 8.49 \\
\bottomrule
\end{tabular*}
\end{table}

\subsection{Impact Statement}
\label{sec:impact}

This paper presents work whose goal is to advance the field of Machine Learning by developing a scalable and analytically tractable Bayesian framework for uncertainty quantification in regression. The proposed Bayesian Deep Kernel Networks (BDKN) are intended to improve the reliability and interpretability of predictive uncertainty relative to commonly used baselines (e.g., Gaussian Processes, Bayesian last-layer models, and deep ensembles), which can be beneficial for downstream decision-making in risk-sensitive settings.

Potential positive societal impacts include enabling more trustworthy model deployment by reducing overconfident predictions, supporting better uncertainty-aware decisions in applications such as scientific modeling, engineering design, and other domains where regression is a core component. By providing a transparent posterior over ensemble-member contributions and closed-form predictive uncertainty, this work may also facilitate diagnostic analyses that help practitioners detect brittle predictors and improve model accountability.

To the best of our knowledge, this work does not introduce direct negative societal impacts or high-risk misuse concerns.

%%%%%%%%%%%%%%%%%%%%%%%%%%%%%%%%%%%%%%%%%%%%%%%%%%%%%%%%%%%%%%%%%%%%%%%%%%%%%%%

% \newpage
% \input{checklist.tex}

\end{document}